\documentclass[conference]{IEEEtran}
\IEEEoverridecommandlockouts
\usepackage{cite}
\usepackage{amsmath,amssymb,amsfonts}
\usepackage{algorithmic}
\usepackage{graphicx}
\usepackage{textcomp}
\usepackage{xcolor}

\usepackage{subfig}
\usepackage{multirow}

\def\BibTeX{{\rm B\kern-.05em{\sc i\kern-.025em b}\kern-.08em
    T\kern-.1667em\lower.7ex\hbox{E}\kern-.125emX}}
\begin{document}

\title{Comparison and Benchmarking of AI Models and Frameworks on Mobile Devices\\
%should not be used}\\
}

%%%%%%%%%%%%% comment out for double blind review %%%%%%%%%%%%%%%%%
 \author{\IEEEauthorblockN{1\textsuperscript{st} Chunjie Luo}
 \IEEEauthorblockA{\textit{Institute of Computing Technology} \\
 \textit{Chinese Academy of Sciences}\\
\textit{BenchCouncil}\\
 Beijing, China \\
 luochunjie@ict.ac.cn}
 \and
\IEEEauthorblockN{2\textsuperscript{nd} Xiwen He}
 \IEEEauthorblockA{\textit{Institute of Computing Technology} \\
 \textit{Chinese Academy of Sciences}\\
\textit{BenchCouncil}\\
Beijing, China\\
hexiwen@ict.ac.cn}
 \and
\IEEEauthorblockN{3\textsuperscript{nd} Jianfeng Zhan \thanks{Jianfeng Zhan is the corresponding author.}}
 \IEEEauthorblockA{\textit{Institute of Computing Technology} \\
 \textit{Chinese Academy of Sciences}\\
\textit{BenchCouncil}\\
Beijing, China\\
zhanjianfeng@ict.ac.cn }
 \and
\IEEEauthorblockN{4\textsuperscript{nd} Lei Wang}
 \IEEEauthorblockA{\textit{Institute of Computing Technology} \\
 \textit{Chinese Academy of Sciences}\\
\textit{BenchCouncil}\\
Beijing, China\\
wanglei\_2011@ict.ac.cn}
 \and
\IEEEauthorblockN{5\textsuperscript{nd} Wanling Gao}
 \IEEEauthorblockA{\textit{Institute of Computing Technology} \\
 \textit{Chinese Academy of Sciences}\\
\textit{BenchCouncil}\\
Beijing, China\\
gaowanling@ict.ac.cn}
 \and
\IEEEauthorblockN{6\textsuperscript{nd} Jiahui Dai}
 \IEEEauthorblockA{\textit{Beijing Academy of Frontier } \\
\textit{Sciences and Technology}\\
Beijing, China\\
daijiahui@mail.bafst.com}

% \and
 }

\maketitle

\begin{abstract}
Due to increasing amounts of data and compute resources,  deep learning achieves many successes in various domains.  The application of deep learning on the mobile and embedded devices is taken more and more attentions, benchmarking and ranking the AI abilities of mobile and embedded devices becomes an urgent problem to be solved. Considering the model diversity and framework diversity, we propose a benchmark suite, AIoTBench, which focuses on the evaluation of the inference abilities of mobile and embedded devices. AIoTBench covers three typical heavy-weight networks: ResNet50, InceptionV3, DenseNet121, as well as three light-weight networks: SqueezeNet, MobileNetV2, MnasNet.  Each network is implemented by three frameworks which are designed for mobile and embedded devices: Tensorflow Lite, Caffe2, Pytorch Mobile. To compare and rank the AI capabilities of the devices, we propose two unified metrics as the AI scores: Valid Images Per Second (VIPS) and Valid FLOPs Per Second (VOPS). Currently, we have compared and ranked 5 mobile devices using our benchmark. This list will be extended and updated soon after.

\end{abstract}

\section{Introduction}
Due to increasing amounts of data and compute resources, deep learning achieves many successes in various domains. To prove the validity of the AI, researchers take more attention to train more accurate models in the early days.  Since many models have achieved pretty performance in various domains, the applications of the AI models to the end-users are put on the agenda.  
%After a model is trained, it is deployed and applied to run inference.  
After entering the application period, the inference becomes more and more important. According to a IDC report \cite{idcreport}, the demand for AI inference will far exceed the demand for training in the future.

With the popularity of smart phones and internet of things, researchers and engineers have made effort to apply the AI inference on the mobile and embedded devices.
Running the inference on the edge devices can 1) reduce the latency,  2) protect the privacy, 3) consume the power \cite{tflitemodel}.

To make the inference on the edge devices more efficient, the neural networks are designed more light-weight by using simpler architecture, or by quantizing, pruning and compressing the networks. Different networks present different trade-offs between accuracy and computational complexity. These networks are designed with very different philosophies and have very different architectures. There is no single network architecture that unifies the network design and application.  Beside of the diversity of models, the AI frameworks on mobile and embedded devices are also diverse, e.g. Tensorflow Lite \cite{tflitemodel}, Caffe2 \cite{caffe2}, Pytorch Mobile \cite{pytorchmobile}. 
On the other hand, the mobile and embedded devices provide additional hardware acceleration using GPUs or NPUs to support the AI applications. 
Since AI applications on mobile and embedded devices get more and more attentions, benchmarking and ranking the AI abilities of those devices becomes an urgent problem to be solved.

MLPerf Inference \cite{reddi2019mlperf} proposes a set of rules and practices to ensure comparability across systems with wildly differing architectures. It defines several high-level abstractions of AI problems, and specifies the task to be accomplished and the general rules of the road, but leaves implementation details to the submitters.
MLPerf Inference aims to more general and flexible benchmarking. However, this flexibility brings unfeasibility and unaffordable for a specific domain. The users of the benchmark need make their efforts to optimize the whole stack of the AI solutions, from the languages, libraries, frameworks, to hardwares.  Moreover, the excessive optimization of a particular solution may affect the generality of the AI system.
ETH Zurich AI Benchmark \cite{ignatov2018ai, Ignatov2019AIBA} aims to evaluate the AI ability of Android smartphones. Although AI Benchmark covers diverse tasks, we believe that the model diversity should be taken more attentions, since the models are often shared between different tasks by transfer learning. For example, the pre-trained networks for image recognition are often used as the backbone to extract the feature maps for other vision tasks, e.g. the object detection and segmentation.  Moreover, AI Benchmark is implemented only using TensorFlow Lite. 

In this paper, we propose a benchmark suite, AIoTBench, which focuses on the evaluation of the inference ability of mobile and embedded devices. The workloads in our benchmark cover more model architectures and more frameworks. 
Specifically, AIoTBench covers three typical heavy-weight networks: ResNet50 \cite{he2016deep}, InceptionV3 \cite{szegedy2016rethinking}, DenseNet121 \cite{huang2017densely}, as well as three light-weight networks: SqueezeNet \cite{iandola2016squeezenet}, MobileNetV2 \cite{sandler2018mobilenetv2}, MnasNet \cite{tan2019mnasnet}.  Each model is implemented  by three frameworks: Tensorflow Lite, Caffe2, Pytorch Mobile.
Comparing to MLPerf Inference, our benchmark is more available and affordable for the users, since it is off the shelf and needs no re-implementation.

Our benchmark can be used for:
\begin{itemize}
\item Comparison of different AI models. Users can make a tradeoff between the accuracy, model complexity, model size and speed depending on the application requirement.
\item Comparison of different AI frameworks on mobile environment. Depending on the selected model, users can make comparisons between the AI frameworks on mobile devices.
\item Benchmarking and ranking the AI abilities of different mobile devices. With diverse and representative models and frameworks, the mobile devices can get a comprehensive benchmarking and evaluation.
\end{itemize}

To compare and rank the AI capabilities of the devices, we propose two unified metrics as the AI scores: Valid Images Per Second (VIPS) and Valid FLOPs Per Second (VOPS). They reflect the trade-off between quality and performance of the AI system.
Currently, we have compared and ranked 5 mobile devices using our benchmark: Galaxy s10e, Honor v20, Vivo x27, Vivo nex,and Oppo R17. This list will be extended and updated soon after.

\section{Related Work}
MLPerf Inference \cite{reddi2019mlperf} proposes a set of rules and practices to ensure comparability across systems with wildly differing architectures. 
Unlike traditional benchmarks, e.g. SPEC CPU, which runs out of the box, MLPerf Inference is a semantic-level benchmark. It defines several high-level abstractions of AI problems, and specifies the task to be accomplished and the general rules of the road, but leaves implementation details to the submitters \footnote{For mobile devices, there is a preliminary reference implementation of MLPerf Inference based on Tensorflow Lite, but it is not official yet.}.
MLPerf Inference aims to more general and flexible benchmarking. For a specific domain, however, this flexibility may bring unfeasibility and unaffordable. The users need make their efforts to optimize the whole stack of the AI solutions, from the languages, libraries, frameworks, to hardwares. For the hardware manufacturer, the implementation and optimization of the AI task may take more proportion to achieve a higher score of the benchmarking.  Moreover, the excessive optimization of a particular solution may affect the generality of the system.

ETH Zurich AI Benchmark \cite{ignatov2018ai, Ignatov2019AIBA} aims to evaluate the AI ability of Android smartphones. It contains several workloads covering the tasks of object recognition, face recognition, playing atari games, image deblurring, image super-resolution, bokeh simulation, semantic segmentation, photo enhancement. Although AI Benchmark covers diverse tasks, we believe that the model diversity should be taken more attentions, since the models are often shared between different tasks by transfer learning. For example, the pre-trained networks for image recognition are often used as the backbone to extract the feature maps for other vision tasks, e.g. object detection and segmentation.  Moreover, AI Benchmark is implemented only using TensorFlow Lite. 

Simone Bianco et al. \cite{bianco2018benchmark} analyzes more than 40 state-of-the-art deep architectures in terms of accuracy rate, model complexity, memory usage, computational complexity, and inference time. They experiment the selected architectures on a workstation equipped with a NVIDIA Titan X Pascal and an embedded system based on a NVIDIA Jetson TX1 board. Their workloads are implemented using PyTorch. 
Shaohuai Shi et al. \cite{shi2016benchmarking} compare the state-of-the-art deep learning software frameworks, including Caffe, CNTK, MXNet, TensorFlow, and Torch. They compare the running performance of these frameworks with three popular types of neural networks on two CPU platforms and three GPU platforms. 
The paper \cite{Luo2018AIoTBT} discusses a comprehensive design of benchmark on mobile and embeded devices, while we focus on the vision domain in this paper. 
%AIBench \cite{gao2019aibench} is an industry-scale end-to-end AI benchmark suite, which abstracts and identifies sixteen prominent AI problem domains. 
Other AI benchmarks include Fathom \cite{adolf2016fathom}, DAWNBench \cite{coleman2017dawnbench}.
These benchmarks are not designed for mobile devices. 

Our benchmark focuses on the evaluation of the inference ablity of mobile and embedded devices. The workloads in our benchmark cover more model architectures and more frameworks. Additionally, comparing to MLPerf Inference, our benchmark is more available and affordable for the users, since it is off the shelf and needs no re-implementation. The details of comparisons of different benchmarks are shown in Table \ref{tab-comp}.

\begin{table*}[htb!]
  \caption{Comparisons between AIoTbench, MLPerf Inference and AI Benchmark}
  \centering
  \begin{tabular}{|c|c|c|c|c|}
    \hline
 &  &  \textbf{AIoTBench}  &  \textbf{MLPerf Inference}  & \textbf{AI Benchmark} \\
    \hline

 \multirow{6}{*}{Model Architectures} & ResNet & \checkmark & \checkmark & \checkmark \\

   &Inception & \checkmark  & $\times$ & \checkmark\\

  & DenseNet & \checkmark & $\times$ & $\times$ \\

  & SqueezeNet & \checkmark & $\times$ &$\times$ \\

   &MobileNet & \checkmark & \checkmark & \checkmark \\

   &MnasNet & \checkmark & $\times$ & $\times$ \\
\hline
\multirow{3}{*}{Implementation Frameworks} & Tensorflow Lite & \checkmark & $\times$ & \checkmark\\
   &  Caffe2 &  \checkmark & $\times$ & $\times$  \\
   &  Pytorch Mobile & \checkmark  &  $\times$  & $\times$\\
\hline
  \end{tabular}
  \label{tab-comp}
\end{table*}

\section{Considerations} \label{sec-considerations}
In this section, we highlight the main considerations of AIoTBench. The details of AIoTBench are described in the next sections.

\subsection{Model Diversity}

\begin{figure}[tbh!]
\centering
\includegraphics[width=1\columnwidth]{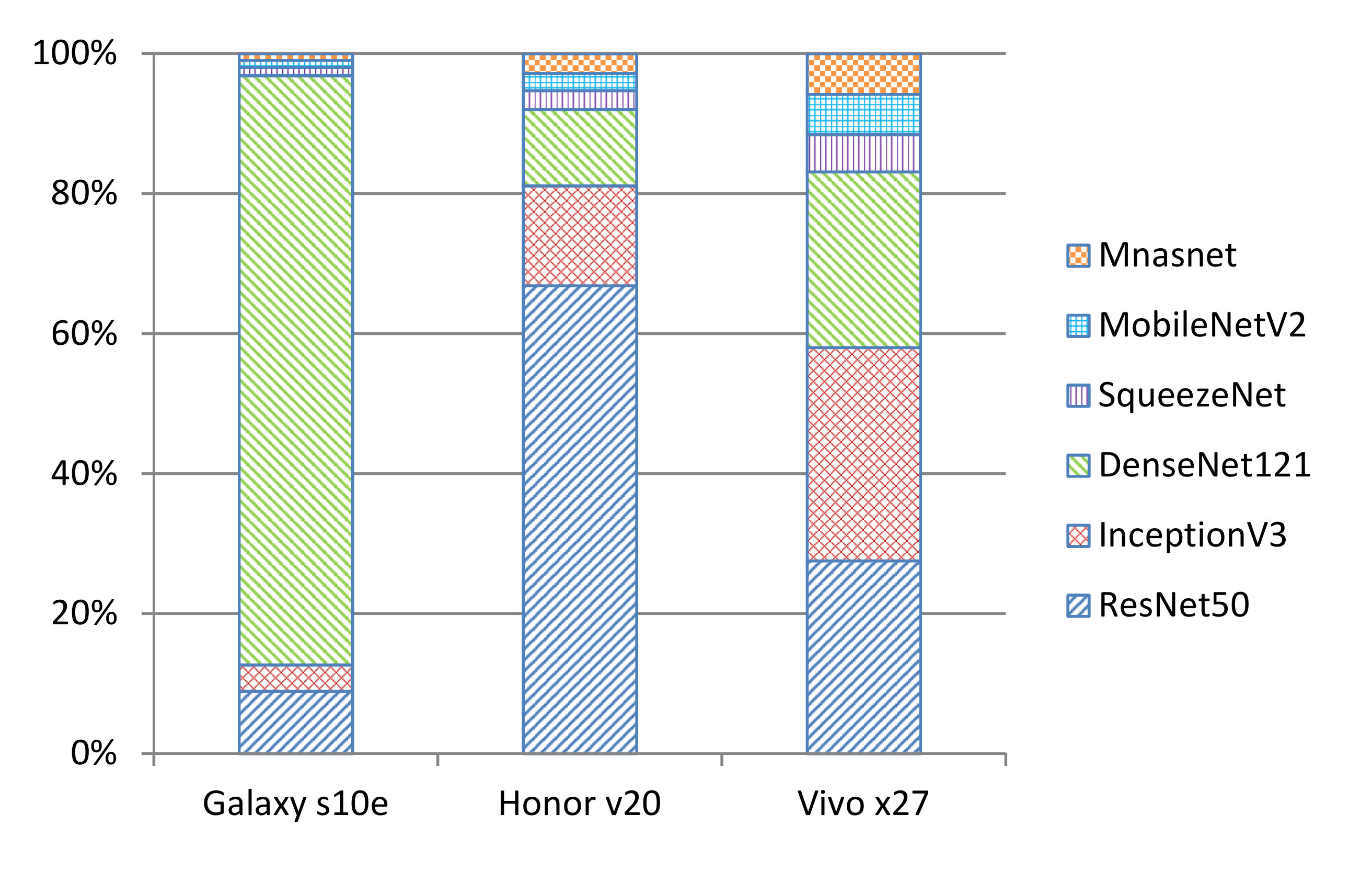}
\caption{The percentage stacked bar of the inference time of diverse models implemented by Tensorflow Lite with NNAPI delegate. Different devices have different degree of support for diverse models.  }
\label{fig-model-diversity}
\end{figure}

Since AlexNet \cite{Krizhevsky2012ImageNetCW} won the ImageNet Large-Scale Visual Recognition Competition (ImageNet-1k) \cite{russakovsky2015imagenet} in 2012, more accurate networks as well as more efficient networks emerge in large numbers. There is no single network architecture that unifies the network design and application. Thus the workloads of AI benchmark should cover representative and diverse network architectures. Moreover, as shown in Fig.~\ref{fig-model-diversity}, different devices have different degree of support for different models. It is necessary to use diverse models for a comprehensive evaluation.

\subsection{Framework Diversity}

\begin{figure}[tbh!]
\centering
\includegraphics[width=1\columnwidth]{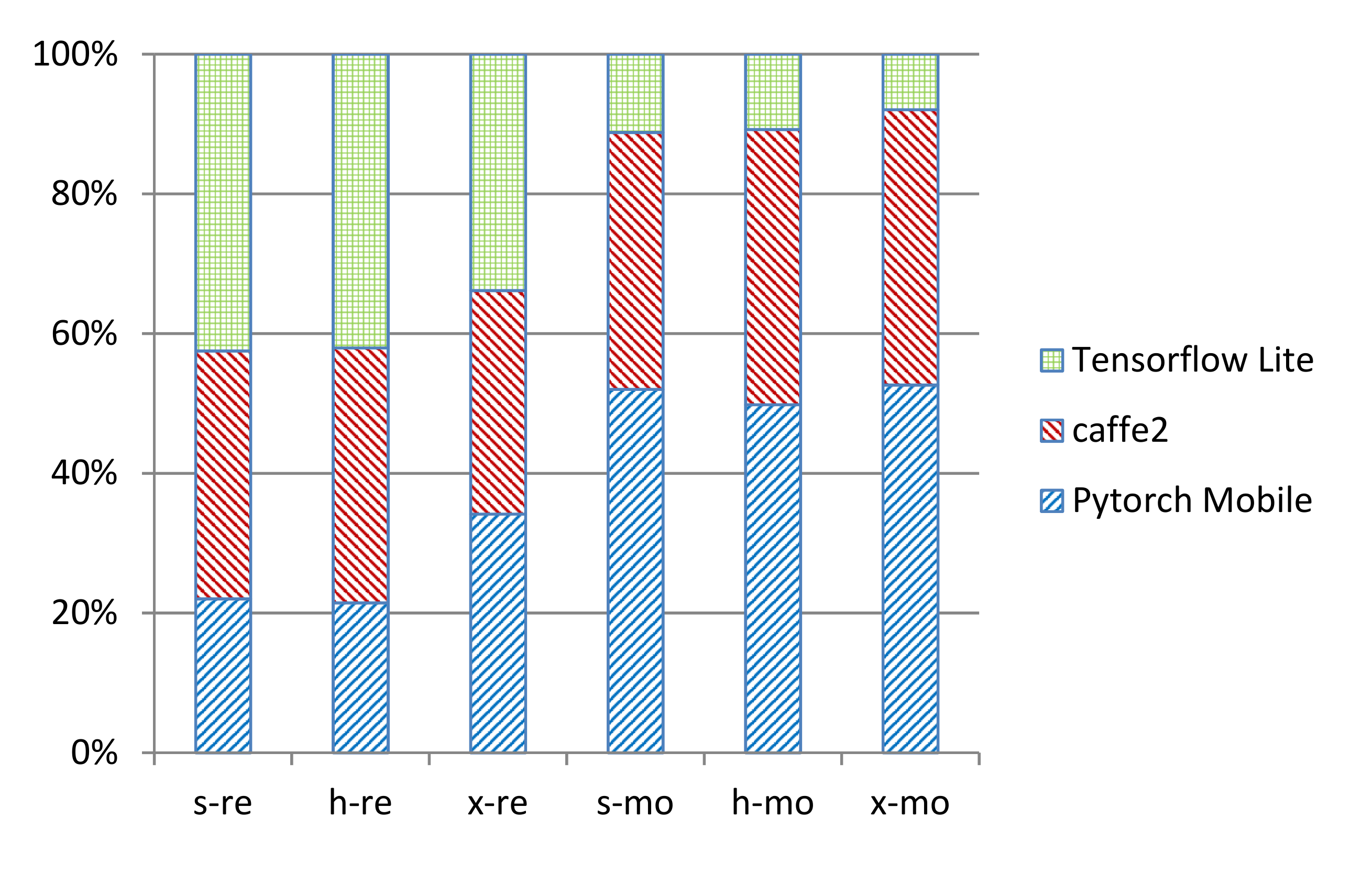}
\caption{The percentage stacked bar of the inference time of diverse frameworks. For different models, different devices have different degree of support for diverse frameworks. (s: Galaxy s10e, h: Honor v20, x: Vivo x27. re: ResNet50, mo: MobileNetV2. ) }
\label{fig-framework-diversity}
\end{figure}

With the growing success of deep learning, there has come out lots of popular open source deep learning frameworks, e.g. Tensorflow, Pytorch, Caffe, CNTK, PaddlePaddle, and MXNet. Each framework has a corporate giant and a wide community behind it. These frameworks are also developing their special solutions for deploying the model on mobile, embedded, and IoT devices. Benchmarks should cover diverse frameworks to reflect this complex ecology. From Fig.~\ref{fig-framework-diversity}, we can see that different devices have different degree of support for diverse frameworks.

\section{Task and Dataset} \label{sec-task}

Currently, AIoTBench focuses on the task of image classification. A classifier network takes an image as input and predicts its class. Image classification is a key task of pattern recognition and artificial Intelligence. It is intensively studied by the academic community, and widely used in commercial applications. The classifier network is also widely used in other vision tasks. It is a common practice that pre-training the classifier network on the ImageNet classification dataset, then replacing the last loss layer and fine-tuning according to the new vision task by transfer learning. 
The pre-trained network is used as the feature extractor in the new vision task.
%and accounts the major computational consumption in the new vision task. 
For example, the object detection and segmentation often use the pre-trained classifier networks as the backbone to extract the feature maps. 

The classifier network is also widely used in other AI benchmarks, and becomes a de facto standard to evaluate AI system.
MLPerf Inference \cite{reddi2019mlperf} includes two image classification tasks: a heavy-weight model ResNet50 and a light-weight model MobileNetV1. It also includes two object detection tasks. Their network architectures are similar with the image classification tasks, a heavy model SSD-ResNet34 using ResNet as the backbone, and a light model SSD-MobileNetV1 using MobileNet as the backbone.
AI Benchmark \cite{ignatov2018ai} includes two image classification tasks, based on MobileNetV1 and InceptionV3, respectively. Its face recognition task uses InceptionV4 architecture as the backbone, image deblurring uses VGG-like architecture, image super-resolution uses VGG-like and ResNet-like architecture, image enhancement use ResNet-like architecture, respectively.

ImageNet 2012 classification dataset \cite{russakovsky2015imagenet} is used in our benchmark. The original dataset has 1280000 training images and 50,000 validation images with 1000 classes.
%We collect models from the model zoo of each framework,  or we collect models trained with other deep learning framework and then we convert them into the target framework. 
%All models in our benchmark are trained on ImageNet 2012 training set, except Inception\_V3 in caffe2, which is trained on ImageNet 2015. 
We focus on the evaluation of the inference of AI models, thus we just need the validation data.
Running the models on the entire validation images on mobile devices takes too long time. So we randomly sample 5 images for each classes from the original validation images. The final validation images we use in our benchmark are 5000 totally.
%and we  use ImageNet val-5k to refer to the final validation data.

\section{Models} \label{sec-model}
\begin{figure*}[tb]
\centering
\subfloat[ResNet50 \cite{he2016deep}]{
\label{res50_arch}
\includegraphics[width=6cm,height=6cm]{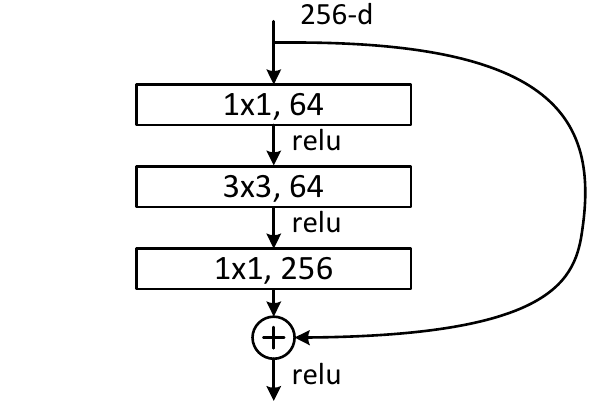}} 
\centering
\subfloat[InceptionV3 \cite{szegedy2016rethinking}]{
\label{inception_arch}
\includegraphics[width=6cm,height=6cm]{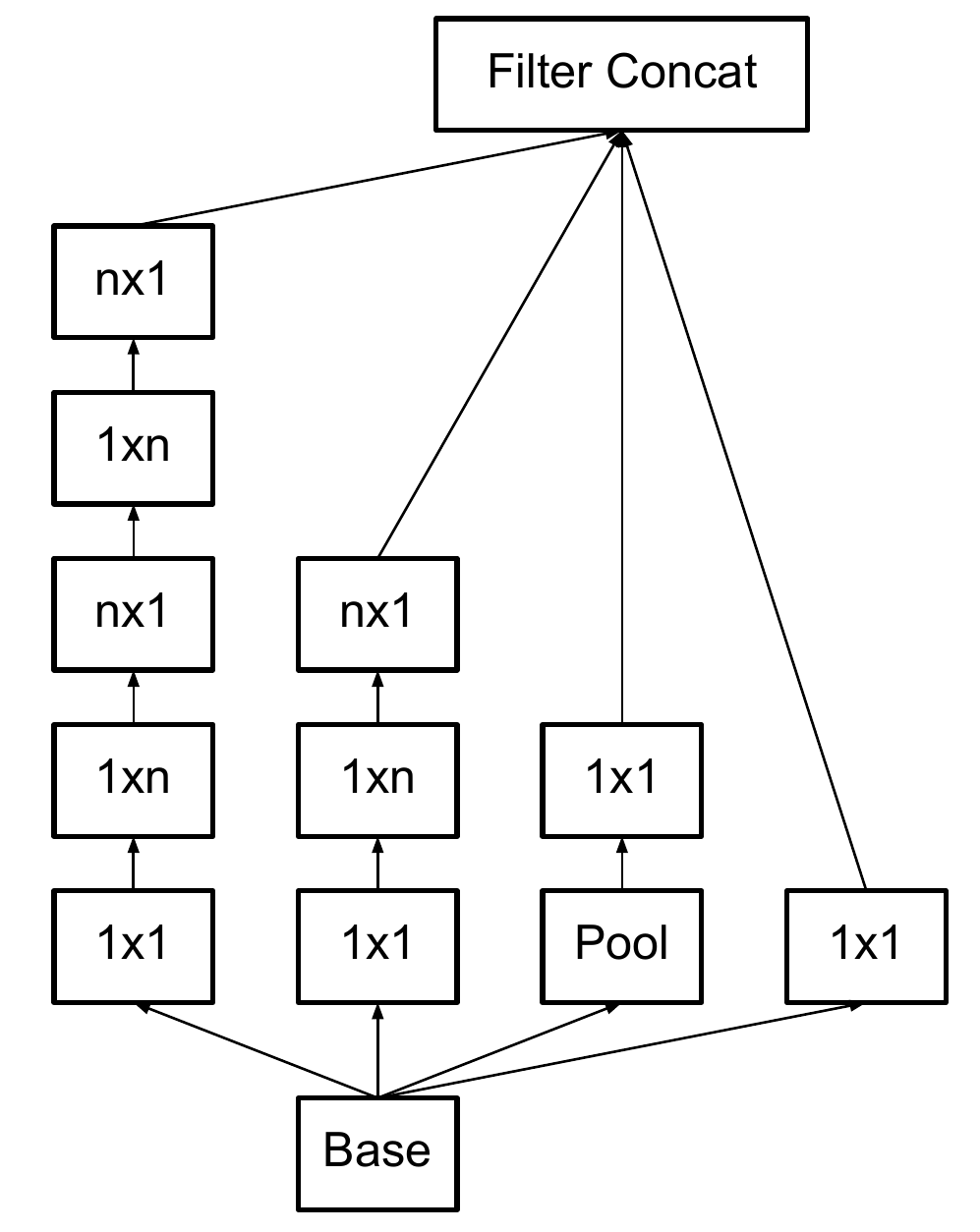}} 
\centering
\subfloat[DenseNet121 \cite{huang2017densely}]{
\label{dense121_arch}
\includegraphics[width=6cm,height=6cm]{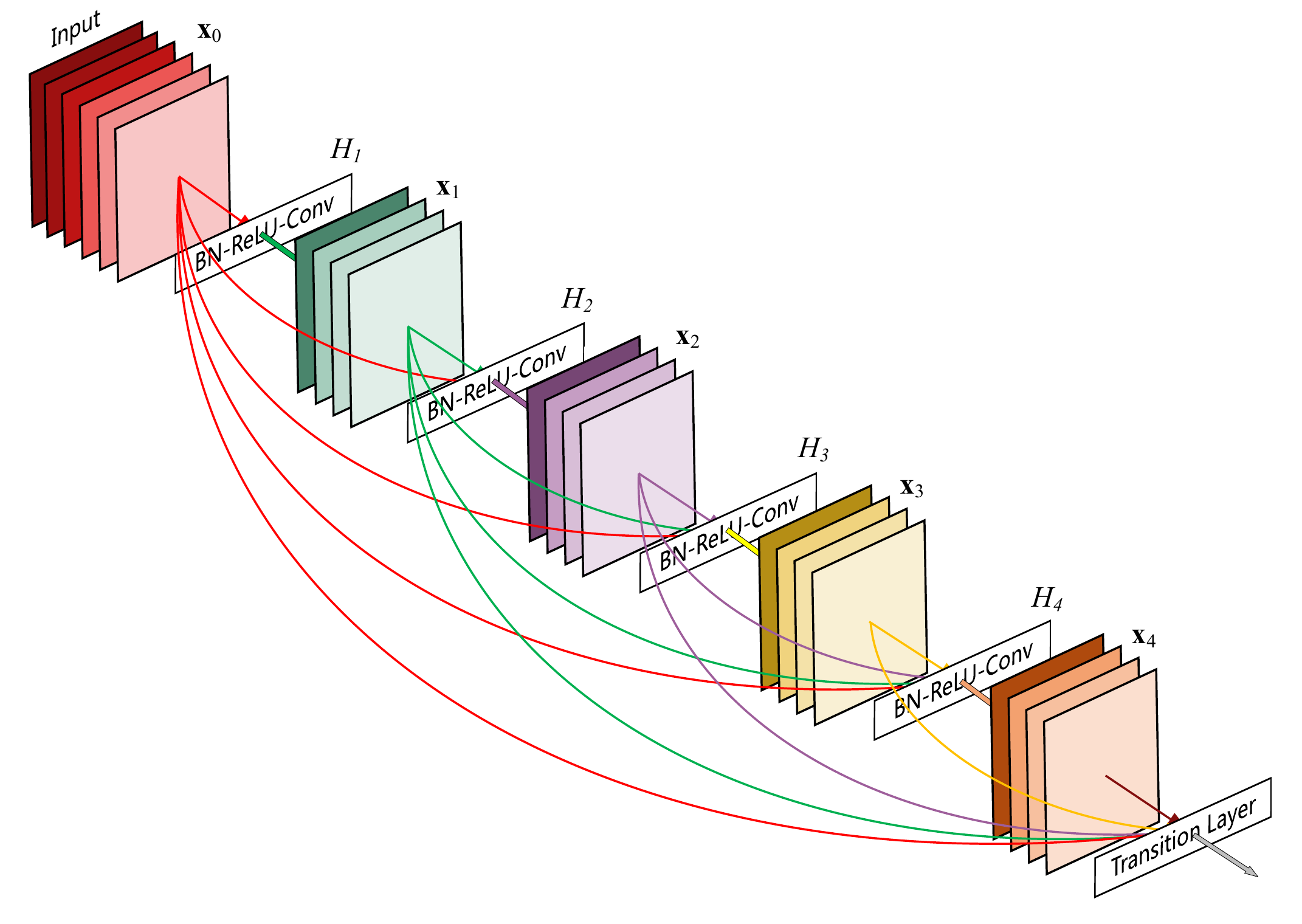}} \\
\centering
\subfloat[SqueezeNet \cite{iandola2016squeezenet}]{
\label{squeeze_arch}
\includegraphics[width=6cm,height=6cm]{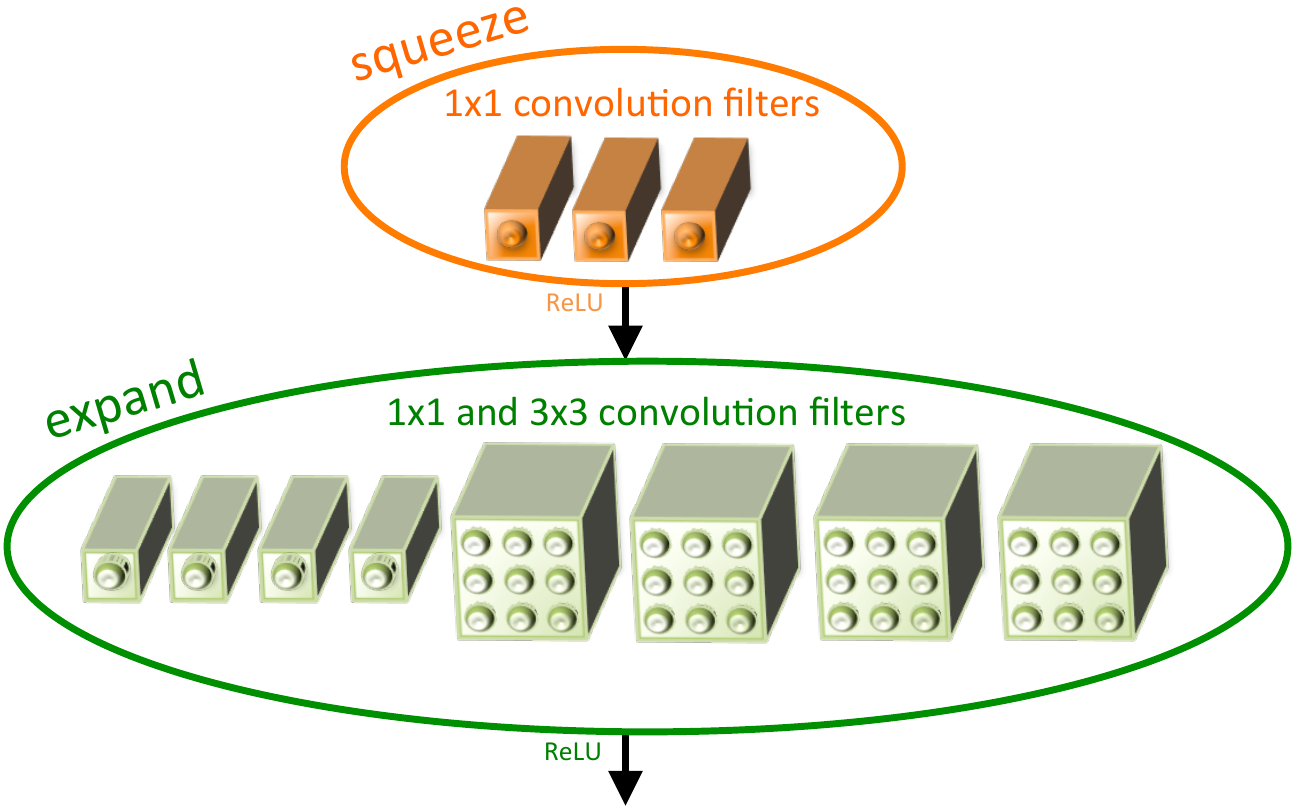}} 
\centering
\subfloat[MobileNetV2 \cite{sandler2018mobilenetv2}]{
\label{mobilenet_arch}
\includegraphics[width=6cm,height=6cm]{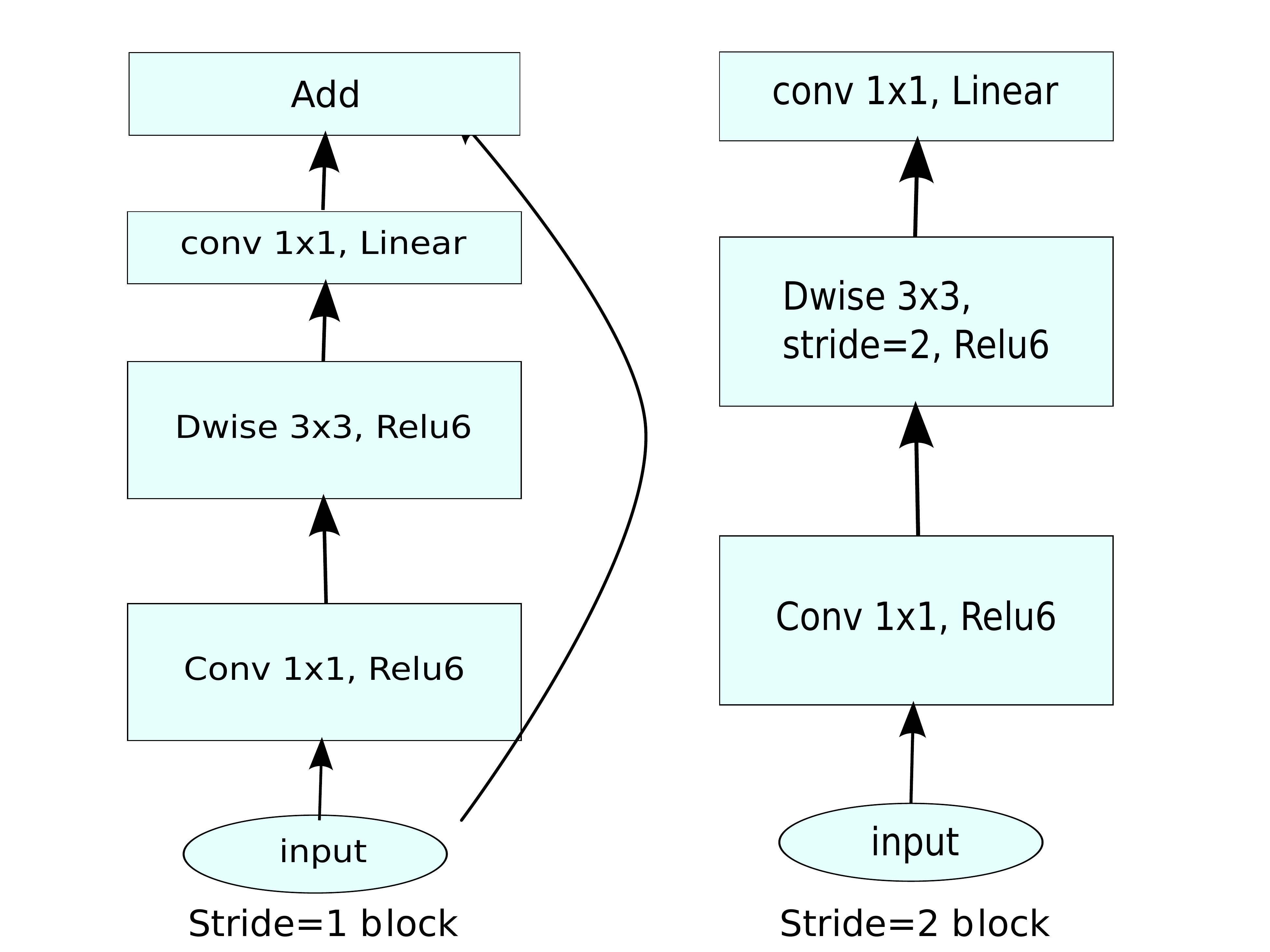}} 
\centering
\subfloat[MnasNet \cite{tan2019mnasnet}]{
\label{mnasnet_arch}
\includegraphics[width=6cm,height=6cm]{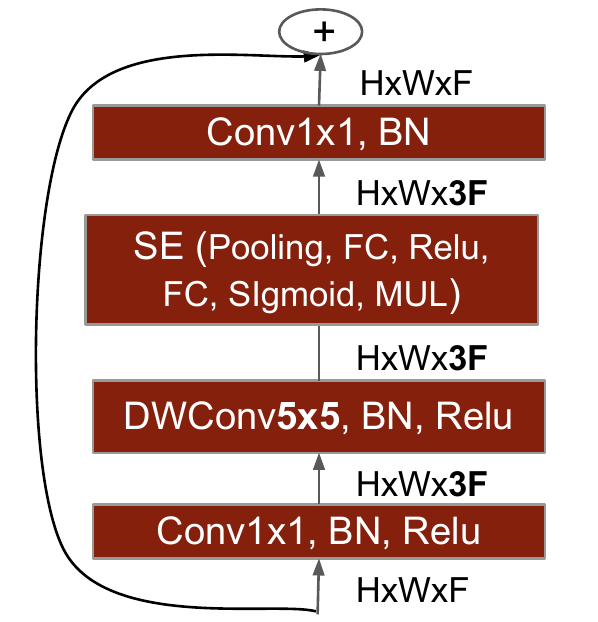}}
\caption{The typical modules of different models}
\label{fig-arch}
\end{figure*}

Although AlexNet plays the role of the pioneer, it is rarely used in modern days. Considering the diversity and popularity, we choose three typical heavy-weight networks: ResNet50 \cite{he2016deep}, InceptionV3 \cite{szegedy2016rethinking}, DenseNet121 \cite{huang2017densely}, as well as three light-weight networks: SqueezeNet \cite{iandola2016squeezenet}, MobileNetV2 \cite{sandler2018mobilenetv2}, MnasNet \cite{tan2019mnasnet}. These networks are designed with very different philosophies and have very different architectures. And they are widely used in the academia and industry. We have also tried the VGG16 \cite{Simonyan2014VeryDC}, which has 138 million parameters and 15300 million FLOPs. It is extremely slow when running on mobile devices. In other words, it is too heavy-weight to be used in mobile devices. The typical modules of the models used in our benchmark are shown in Fig.~\ref{fig-arch}. And Table \ref{diffmodel} presents the FLOPs \footnote{FLOPs (some literatures use Mult-Adds) refers to the number of multiply-accumulates  to compute the model inference on a single image. FLOPs is widely adopted as the metric to reflect the computational complexity of a model.}, parameters, and the original reference accuracy of the selected models.  

\begin{table}[htb!]
  \caption{The FLOPs, parameters, and the original reference accuracy of the models used in  AIoTBench}
  \centering
  \begin{tabular}{|c|c|c|c|}
    \hline
    \multirow{2}{*}{\textbf{Model}} & \textbf{FLOPs}  &  \textbf{Parameters}  & \textbf{Accuracy} \\
     &  (Millions)  & (Millions) &  (\%)\\
    \hline

   ResNet50 & 3800 & 25.6 & 76 \\

   InceptionV3 & 5000  & 23.2 & 78.8\\

   DenseNet121 & 2800 & 8  &  74 \\

   SqueezeNet & 833 & 1.25 & 57.5 \\

   MobileNetV2 & 300 & 3.4 & 72 \\

   MnasNet & 315 & 3.9 & 75.2 \\
\hline
  \end{tabular}
  \label{diffmodel}
\end{table}

\subsection{ResNet}
ResNet \cite{he2016deep} was proposed by Microsoft Research in 2015. In ILSVRC (ImageNet Large-Scale Visual Recognition Challenge \cite{russakovsky2015imagenet} ) 2015 and Microsoft COCO (Microsoft Common Objects in Context \cite{Lin2014MicrosoftCC} ) 2015 competitions, methods based on ResNet won the 1st places on the tasks of ImageNet classification, ImageNet detection, ImageNet localization, COCO detection, and COCO segmentation. Moreover, it won the best paper award of CVPR 2016 \cite{cvpr2016}. 
Many variants of ResNet have proposed later, e.g. PreAct ResNet \cite{He2016IdentityMI}, ResNeXt\cite{Xie2017AggregatedRT}. Considering the tradeoff between accuracy and computation cost, ResNet50 is the most popular architecture among the family of ResNet and its variants. Fig.~\ref{res50_arch} shows the typical module of ResNet50. It can be defined as $y=F(x)+x$, where $x$ and $y$ are the input and output feature maps. The function $F$ consists of 1x1, 3x3 and 1x1 convolutional layers. The key contribution is the identity mapping ($+x$) which is performed by a shortcut
connection and element-wise addition. Without extra parameter and computation complexity, identity mapping significantly alleviates the vanishing gradient problem which hinders the training of very deep networks. With 25.6 million parameters and 3800 million FLOPs, ResNet50 achieves 76\% validation accuracy on ImageNet.

\subsection{Inception}
The first version of Inception, also called GoogLeNet \cite{Szegedy2015GoingDW},  was proposed by Google in 2014.
GoogLeNet is the winner of the task of classification and detection in the ILSVRC 2014. GoogLeNet increases the depth and width of the network while keeping the computational budget constant. It uses filter sizes of 1x1, 3x3 and 5x5 in different branches to capture multi-scale information of the feature.
The computational efficiency and practicality are the main design considerations of Inception architecture. InceptionV2 and InceptionV3 \cite{Szegedy2015RethinkingTI} have a similar architecture. As shown in Fig.~\ref{inception_arch}, InceptionV3 factorizes convolutions with large filter size into smaller convolutions, e.g. using two stacked 3x3 convolutional filters instead of 5x5 filter. Further, it spatially factorizes standard convolutions into asymmetric convolutions, e.g. using 1x3 following by 3x1 instead of 3x3. Comparing to InceptionV2, InceptionV3 uses several training tricks to achieve higher accuracy.  InceptionV4 \cite{Szegedy2016Inceptionv4IA} adds residual connection into Inception architecture. We choose InceptionV3 in our benchmark because of its extensive use in the community. InceptionV3 have 23.2 million parameters and 5000 million FLOPs, and it achieves 78.8\% accuracy.

\subsection{DenseNet}
DenseNet \cite{Huang2016DenselyCC} was proposed by Cornell University, Tsinghua University and Facebook in 2017. It won the best paper award of CVPR 2017 \cite{cvpr2017}. Fig.~\ref{dense121_arch} shows the dense block of DenseNet. Different from ResNet which use identity mapping to make shortcut connection,  DenseNet connects all layers directly with each other by concatenating them. Each layer concatenates all its preceding layers and passes on itself to all subsequent layers. By direct and dense connection, DenseNet alleviates the vanishing gradient problem and strengthen feature propagation. The densely connected block use very narrow layers, thus reduce the number of parameters. Although each output layer is narrow, the dense block typically has many more inputs since it concatenates all its preceding layers. To reduce the number of input and improve computational efficiency, a 1x1 convolution is used as bottleneck before each 3x3 convolution. DenseNet121 has 8 million parameters, 2800 million FLOPs, and achieve 74\% accuracy.

\subsection{SqueezeNet}
SqueezeNet \cite{iandola2016squeezenet} was proposed by UC Berkeley and Stanford University in 2016. 
The main purpose of SqueezeNet is to decrease the number of parameters while maintaining competitive accuracy. SqueezeNet employs three main strategies when designing CNN architectures to decrease the parameters: 1) Replace 3x3 filters with 1x1 filters, 2) Decrease the number of input channels to 3x3 filters, 3) Downsample late in the network so that convolution layers have large activation maps. Following these strategies, the Fire module, as shown in Fig.~\ref{squeeze_arch}, is designed: 1) a squeeze convolution layer which has only 1x1 filters, 2) followed by an expand layer that has a mix of 1x1 and 3x3 convolution filters. SqueezeNet achieves AlexNet-level accuracy on ImageNet with 50x fewer parameters. It has only 1.25 million of total parameters, but achieves 57.5\% accuracy.
Since SqueezeNet focuses on the compression of model size, it dose not consider the computation complexity when designing, and have 833 million FLOPs. 

\subsection{MobileNet}
MobileNet is another series of models proposed by Google. As a light-weight deep neural network,  MobileNet is designed and optimized for mobile and embedded vision applications. The first version of MobileNet \cite{Howard2017MobileNetsEC} is published in 2016, the second version MobileNetV2 \cite{Sandler2018MobileNetV2IR} is published in 2018, and the third version MobileNetV3 \cite{Howard2019SearchingFM} is published in 2019. 
MobileNet is based on depthwise separable convolutions to reduce the number of parameters and computation FLOPs.
Depthwise separable convolutions, initially proposed in \cite{Sifre2014RigidMotionSF}, factorizes a standard convolution into a depthwise convolution and a pointwise convolution. 
The depthwise convolution applies a single 3x3 filter to each input channel to capture the spatial relationship of features. The pointwise convolution applies a 1x1 convolution to capture the channel-wise relationship of features. 
This factorization is widely adopted by the following design of light-weight neural networks, e.g. MobileNetV2 \cite{Sandler2018MobileNetV2IR}, MobileNetV3 \cite{Howard2019SearchingFM}, ShuffleNet \cite{Zhang2017ShuffleNetAE}, ShuffleNetV2 \cite{Ma2018ShuffleNetVP}, MnasNet \cite{Tan2018MnasNetPN}. Beside depthwise separable convolutions,  MobileNetV2 introduce two optimized mechanisms: 1) inverted residual structure where the shortcut connections are between the thin layers. 2) linear bottlenecks which removes non-linearities in the narrow layers. MobileNetV3 is searched by hardware-aware network architecture search (NAS) algorithm. Considering the popularity and diversity, we choose the MobileNetV2, and another light-weight architecture searched by NAS, MnasNet, which is described in Section \ref{mnasnet}.  MobileNetV2 has 3.4 million parameters, 300 million FLOPs, and achieve 72\% accuracy.

\subsection{MnasNet} \label{mnasnet}
Automated machine learning (AutoML) has emerged as a hot topic. As the main domain of AutoML, neural architecture search (NAS) \cite{Zoph2016NeuralAS, Baker2016DesigningNN, Liu2017ProgressiveNA, Pham2018EfficientNA} automatically find the architecture in the pre-defined search space. It can find the optimal combination structure of existing neural unit, but can not invent new techniques. NAS reduces the demand for experienced human experts comparing to hand-drafted design. 
MnasNet \cite{Tan2018MnasNetPN} is a neural architecture automated searched for mobile device by using multi-objective optimization and factorized hierarchical search space. It is proposed by Google in 2018. Instead of using FLOPs to approximate inference latency, the real-world latency is directly measured by executing the model on real mobile devices. In fact, the final architecture of MnasNet is very heterogeneous. Fig.~\ref{mnasnet_arch} shows one of the typical modules of MnasNet.  The module is sequence of 1x1 pointwise convolution, 3x3 depthwise convolution, Squeeze-and-Excitation module (Squeeze-and-Excitation is a light-weight attention module originally proposed in \cite{Hu2019SqueezeandExcitationN}, which is the winner of the  ILSVRC 2017 classification task.), and 1x1 pointwise convolution. MnasNet has 3.9 million parameters, 315 million FLOPs, and achieve 75.2\% accuracy.

\section{Frameworks}

For the mobile and embedded devices, the framework, with which the models are implemented, is also part of the workload. In AIoTBench, each model is implemented by three frameworks: Tensorflow Lite, Caffe2, Pytorch Mobile.

\subsection{Tensorflow Lite}
TensorFlow \cite{tensorflow, Abadi2016TensorFlowAS}, released by Google, is a free and open-source software library for dataflow and differentiable programming. It is widely used for machine learning applications such as neural networks. Tensorflow Lite \cite{tflitemodel} is released in 2017 for deploying the models trained by Tensorflow on mobile, embedded, and IoT devices. It aims to conduct on-device machine learning inference with low latency and a small binary size.  After the model is trained,  it need be converted to  a Tensorflow Lite FlatBuffer file (.lite), and then executed on mobile or embedded devices using the Tensorflow Lite Interpreter. TensorFlow Lite also supports to delegate part or all of graph execution using GPU and NNAPI. It enables on-device machine learning inference with low latency and a small binary size.  
Currently, Tensorflow Lite offers the richest functionality comparing to other AI frameworks on mobile and embedded devices. Moreover, it supports various platforms, e.g. iOS, Android, embedded Linux like Raspberry Pi or Arm64-based boards, and Microcontrollers.
Tensorflow Lite provides the JAVA and C++ API for the development on mobile and embedded devices. 

\subsection{Caffe2}
Caffe \cite{Jia2014CaffeCA} is an open-source deep learning framework, originally developed at UC Berkeley. Caffe2 \cite{caffe2},  built on the original Caffe and released by Facebook in 2017, is a light-weight and modular framework for production-ready training and deployment. 
In fact, Caffe2 can be used in the cloud or on mobile with its cross-platform libraries.
It also supports various mobile and embeded platforms, e.g. iOS, Android, Raspbian, Tegra.
The Caffe2 model files consist of two parts: a set of weights that represent the learned parameters, which is stored in a init\_net.pb file, and a set of operations that form a computation graph, which is stored in a predict\_net.pb file.
Caffe2 provide the C++ API for the development on mobile and embedded devices.

\subsection{Pytorch Mobile}
Pytorch \cite{pytorch}, primarily developed by Facebook's AI Research lab (FAIR),  is an open-source machine learning library based on the Torch library. It aims to replace for NumPy to use the power of GPUs and provide a deep learning research platform. In 2019, PyTorch Mobile is released within the version of Pytorch 1.3. It supports the deployments of the models on iOS and Android. 
To deploy the model on mobile devices, you need serialize the pre-trained model into a .pt file. 
And then users can use the PyTorch Android API or Pytorch C++ front-end APIs to develop the AI application on mobile devices.

\section{Technical Description}

\subsection{Model sources}
We collect models from the model zoo of each framework, or we collect models trained with other deep learning framework and then we convert them into the target framework.  
\begin{itemize}
\item ResNet50 and DenseNet121 in Tensorflow Lite are converted from the pre-trained models in Keras Applications \cite{kerasapp}.  Other Tensorflow Lite models are collected from the official site of Tensorflow Lite \cite{tflitemodel}, where the models are already converted to the .tflite format.
\item For Caffe2, ResNet50, DenseNet121, SqueezeNet, and MobilenetV2 are directly download from the Caffe2 model zoo \cite{caffe2model}. MnasNet is converted from the pre-trained models in Pytorch Torchvision \cite{torchvision}.  InceptionV3 is converted from the pre-trained Caffe model in \cite{caffemodel}.  
\item All the models in Pytorch Mobile are converted from the pre-trained models in Pytorch Torchvision.
\end{itemize}

\begin{table*}[htb!]
  \caption{The source and model file size of the different model implementation}
  \centering
  \begin{tabular}{|c|c|c|c|}
    \hline
    \textbf{Framework} & \textbf{Model}  &  \textbf{Source}  & \textbf{Model File Size} \\
    \hline

    \multirow{6}{*}{Tensorflow Lite} & ResNet50 & Converted from Keras Application& 98 MB \\
    & InceptionV3 & Tensorflow lite model zoo  & 91 MB\\
    &DenseNet121 & Converted from Keras  Application&  31 MB\\
    &SqueezeNet & Tensorflow lite model zoo  & 4.8 MB \\
   &MobileNetV2 & Tensorflow lite model zoo  &  14 MB\\
   &MnasNet & Tensorflow lite model zoo  & 17 MB \\
    \hline
    \multirow{6}{*}{Caffe2} & ResNet50 & Caffe2 model zoo & 123 MB \\
    & InceptionV3 & Converted from Caffe model  & 115 MB\\
    &DenseNet121 & Caffe2 model zoo  &  39 MB\\
    &SqueezeNet & Caffe2 model zoo & 5.9 MB \\
   &MobileNetV2 & Caffe2 model zoo  &  17 MB\\
   &MnasNet & Converted from Pytorch Torchvision & 22 MB \\
    \hline
    \multirow{6}{*}{Pytorch Mobile} & ResNet50 & Converted from Pytorch Torchvision & 98 MB \\
    & InceptionV3 & Converted from Pytorch Torchvision  & 105 MB\\
    &DenseNet121 & Converted from Pytorch Torchvision  &  32 MB\\
    &SqueezeNet & Converted from Pytorch Torchvision & 4.9 MB \\
   &MobileNetV2 & Converted from Pytorch Torchvision  &  14 MB\\
   &MnasNet & Converted from Pytorch Torchvision  & 18 MB \\
    \hline
  \end{tabular}
  \label{modelsource}
\end{table*}

All models in our benchmark are trained on ImageNet 2012 training set, except InceptionV3 in caffe2, which is trained on ImageNet 2015. When counting the accuracy on the validation data set of ImageNet 2012,  we map the label of ImageNet 2015 to ImageNet 2012.  The same model in Tensorflow Lite and Pytorch Mobile have close file size,  which are smaller than the corresponding one in Caffe2.  The details of the source and model file size are shown in Table \ref{modelsource}.

\subsection{Data preprocessing}
Depending on the training procedure, the input image should be preprocessed into corresponding form for ingestion into pre-trained models. Because of the diverse sources, the data preprocessing of each model of each framework are different from each other. The data preprocessing mainly involves crop and resize, channel arrangement, normalization. 

The images in validation data set are 3 channel RGB (Red-Green-Blue) images of shape (3 x H x W), where H and W are height and width. The input sizes of the models used in our benchmark are 224x224 (except the InceptionV3 whose input size is 299x299). The images should be resized into the corresponding size before ingestion into the model. In practice, there are three common ways to resize the image to the required size:

\begin{enumerate}
\item Firstly, centerly crop the image according to the shortest side (e.g. 360x480 to 360x360). Then resize the image to the model's required size (360x360 to 224x224).
\item  Firstly, resize the image to the shortest side that matches with the model's required size (e.g. 360x 480 to 224x298) while locking down the aspect ratio.  Then, centerly crop the image into final input size (e.g. 224x298 to 224x224).
\item  Firstly, resize the image to a certain size (e.g. 360x 480 to 256x256).  Then, centerly crop the image into final input size (e.g. 256x256 to 224x224).
\end{enumerate}

We experiment the three methods and find that there is little influence on the performance. We choose the first way to preprocess the input image in our benchmark.

Comparing to the data resize, the accuracy is influenced greatly by the channel arrangement and the normalization. Some models need the input shape as (3xHxW), others need (HxWx3). And some models need the channel order as RGB, and others need BGR. 
Before ingested into the model, the input is always normalized. For the image input $x$ which ranges from 0 to 255, different normalizations can be unified as $(x-\mu)/\sigma$, but with different normalization numbers of $\mu$ and $\sigma$.  Unfortunately, the exact training details for some models are unknown. The correct settings are searched through repeatedly experiments. The final settings of the data preprocessing for each model are shown in Table \ref{datapreprocess}.

\begin{table*}[htb!]
  \caption{The data preprocessing of different model implementation. For the image input $x$ which ranges from 0 to 255,  normalization is defined as $(x-\mu)/\sigma$}
  \centering
  \begin{tabular}{|c|c|c|c|c|c|}
    \hline
    \textbf{Framework} & \textbf{Model}  &  \textbf{Input shape} &\textbf{Input type}  & \textbf{Normalization $\mu$} &  \textbf{Normalization $\sigma$}\\
    \hline

    \multirow{6}{*}{Tensorflow Lite} & ResNet50 &  224x224x3 & BGR & (103.939,116.779,123.68) & (1.0,1.0,1.0) \\
    & InceptionV3 & 299x299x3 & RGB  &(127.5,127.5,127.5) & (127.5,127.5,127.5) \\
    &DenseNet121 & 224x224x3 & RGB & (123.675,116.28,103.53)  & (58.395,57.12,57.375) \\
    &SqueezeNet & 224x224x3 & RGB  & (127.5,127.5,127.5) & (255.0,255.0,255.0)  \\
   &MobileNetV2 & 224x224x3 & RGB  &  (127.5,127.5,127.5) &(127.5,127.5,127.5) \\
   &MnasNet & 224x224x3 & RGB &(127.5,127.5,127.5) & (127.5,127.5,127.5) \\
    \hline
    \multirow{6}{*}{Caffe2} & ResNet50 & 3x224x224 & BGR & (127.5,127.5,127.5) & (127.5,127.5,127.5)  \\
    & InceptionV3 & 3x299x299 & BGR & (127.5,127.5,127.5) & (127.5,127.5,127.5)  \\
    &DenseNet121 & 3x224x224 & RGB  & (123.675,116.28,103.53)  & (58.395,57.12,57.375) \\
    &SqueezeNet & 3x224x224 & BGR & (127.5,127.5,127.5)  & (1.0,1.0,1.0)\\
   &MobileNetV2 & 3x224x224 & BGR  & (103.53,116.28,123.675) & (57.375,57.12,58.395)\\
   &MnasNet & 3x224x224 & RGB & (123.675,116.28,103.53) & (58.395,57.12,57.375)  \\
    \hline
    \multirow{6}{*}{Pytorch Mobile} & ResNet50 & 3x224x224 & RGB & (123.675,116.28,103.53) & (58.395,57.12,57.375) \\
    & InceptionV3 & 3x299x299 & RGB  & (123.675,116.28,103.53) & (58.395,57.12,57.375) \\
    &DenseNet121 & 3x224x224 & RGB  & (123.675,116.28,103.53)  & (58.395,57.12,57.375)  \\
    &SqueezeNet & 3x224x224 & RGB &(123.675,116.28,103.53) & (58.395,57.12,57.375)  \\
   &MobileNetV2 & 3x224x224 & RGB & (123.675,116.28,103.53) & (58.395,57.12,57.375) \\
   &MnasNet & 3x224x224 & RGB  & (123.675,116.28,103.53) &  (58.395,57.12,57.375)  \\
    \hline
  \end{tabular}
  \label{datapreprocess}
\end{table*}

\subsection{Test procedure}
Before the test, we need download the model files and validation data into the mobile device.
When testing, the model is first loaded into memory, and then infers the 5000 validation images  sequentially. 
The tests includes 6 models: ResNet50 (using re for short), InceptionV3 (in), DenseNet121 (de), SqueezeNet (sq), MobileNetV2 (mo), MnasNet (mn).
For each model, we test the implementation of  Pytorch Mobile (py), Caffe2 (ca), Tensorflow Lite with CPU (tfc), and Tensorflow Lite with NNAPI delegate (tfn). 
%In the following , we use a short name for a particular test. For example, py-re refer to the test of Pytorch Mobile implementation of ResNet50. 
As a result, we have 24 (6 models * 4 implementations) tests for each device.  The Garbage Collection of JVM and the heat generation of the mobile device may affect the test. For a fair comparison, we shutdown the device and wait at least five minutes of cooling time between each test. The accuracy and the average inference time are logged for counting the final AI score of the measured device.

\subsection{AI Score}
Distilling the AI capabilities of the system to a unified score enables a direct comparison and ranking of different devices. We propose two unified metrics as the AI scores: Valid Images Per Second (VIPS) and Valid FLOPs Per Second (VOPS).

\begin{equation} \label{score_eq}
VIPS = \sum_{i=1}^{n} accuracy_{i} * \frac{1}{time_{i}}
\end{equation}

\begin{equation} \label{score_eq2}
VOPS = \sum_{i=1}^{n} accuracy_{i} * FLOPs_{i} * \frac{1}{time_{i}}
\end{equation}
where $accuracy_i$ refers to the validation accuracy of the $i_{th}$ test, $time_{i}$ refers to the average running time per image of the $i_{th}$ test, and $FLOPs_{i}$ refer to the FLOPs of the model used in the $i_{th}$ test.

The inverse proportion to the average running time ($\frac{1}{time_{i}}$) reflects the throughput of the system. Since the trade-off between quality and performance is always the consideration in AI domain, the accuracy is used as the weight coefficient to compute the final AI score. VIPS is a user-level or application-level metric, since how many images can be processed is the end-user's concern. VOPS is a system-level metric, and it reflects the valid computation that the system can process per second.

The accuracy is an important factor when benchmarking the AI system, since many architectures can trade model quality for lower latency or greater throughput \cite{reddi2019mlperf} \cite{Ignatov2019AIBA}. MLPerf Inference requires that all benchmarking submissions should 
achieve a target accuracy, then comparing different systems using metrics like latency or throughput. MLPerf Inference does not offer a unified  AI score.
Different from MLPerf Inference, AI Benchmark offers a unified AI score. It considers the accuracies as part of over 50 different attributes. The final AI score is calculated as a weighted sum of those attributes. The weight coefficients are calibrated based on the results of particular device. 
We also offer the unified AI scores. When computing the final scores, the accuracies are directly used as the weight coefficients of throughput. Our metrics are more intuitive and explicable. They can be explained as the valid throughput of the system, and reflects the trade-off between quality and performance.

\section{Results}

Currently, we have compared and ranked 5 mobile devices using our benchmark: Samsung Galaxy s10e, Huawei Honor v20, Vivo x27, Vivo nex,and Oppo R17. This list will be extended and updated soon after.

Galaxy s10e, Vivo x27, Vivo nex, and Oppo R17 equip the mobile SoC of Qualcomm SnapDragon.
SnapDragon uses a heterogeneous computing architecture to accelerate the AI applications.
AI Engine of SnapDragon consists of Kryo CPU cores, Adreno GPU and Hexagon DSP.
Honor v20 equips HiSilicon Kirin SoC.
Different from SnapDragon, Kirin introduces a specialized neural processing unit (NPU) to accelerate the AI applications. The detailed configurations of the devices are shown in Table \ref{table-device}. 
%The accuracy, the average inference time and the final AI scores are shown in Table \ref{table-result}.

\begin{table*}[tbh!]
\caption{The features of the measured devices }
\centering
\begin{tabular}{|c|c|c|c|c|c|c|}
\hline
\textbf{Device} &  \textbf{Soc} & \textbf{CPU}  & \textbf{Process} & \textbf{AI Accelerator}  & \textbf{RAM} & \textbf{Android} \\
\hline
\multirow{2}{*}{Galaxy s10e} & \multirow{2}{*}{Snapdragon 855} &  \multirow{2}{*}{Kryo 485, 2.84 GHz}  & \multirow{2}{*}{7nm} & Adreno 640 GPU  & \multirow{2}{*}{6GB}  & \multirow{2}{*}{9} \\
& & & & Hexagon 690 DSP & & \\
\hline
Honor v20  & Kirin 980 & Cortex-A76, 2.6 GHz & 7nm & Cambricon NPU  & 8GB & 9 \\
\hline
\multirow{2}{*}{Vivo x27}  &  \multirow{2}{*}{Snapdragon 710}  & \multirow{2}{*}{Kryo 360, 2.2 GHz} & \multirow{2}{*}{10nm} &  Adreno 616 GPU &  \multirow{2}{*}{8GB}  & \multirow{2}{*}{9}\\
& & & & Hexagon 685 DSP & & \\
\hline
\multirow{2}{*}{Vivo nex}  &  \multirow{2}{*}{Snapdragon 710} & \multirow{2}{*}{Kryo 360, 2.2 GHz} & \multirow{2}{*}{10nm} & \multirow{2}{*}{Adreno 616 GPU} & \multirow{2}{*}{8GB}  & \multirow{2}{*}{9}\\
& & & & Hexagon 685 DSP & & \\
\hline
\multirow{2}{*}{Oppo R17}  &  \multirow{2}{*}{Snapdragon 670} &\multirow{2}{*}{ Kryo 360, 2.0 GHz} & \multirow{2}{*}{10nm} & \multirow{2}{*}{ Adreno 615 GPU} & \multirow{2}{*}{6GB}  & \multirow{2}{*}{8.1}\\
& & & & Hexagon 685 DSP & & \\
\hline
  \end{tabular}
  \label{table-device}
\end{table*}

\textbf{Remark 1: The same AI model on different platform has different accuracy.}

\begin{figure}[tbh!]
\centering
\includegraphics[width=1\columnwidth]{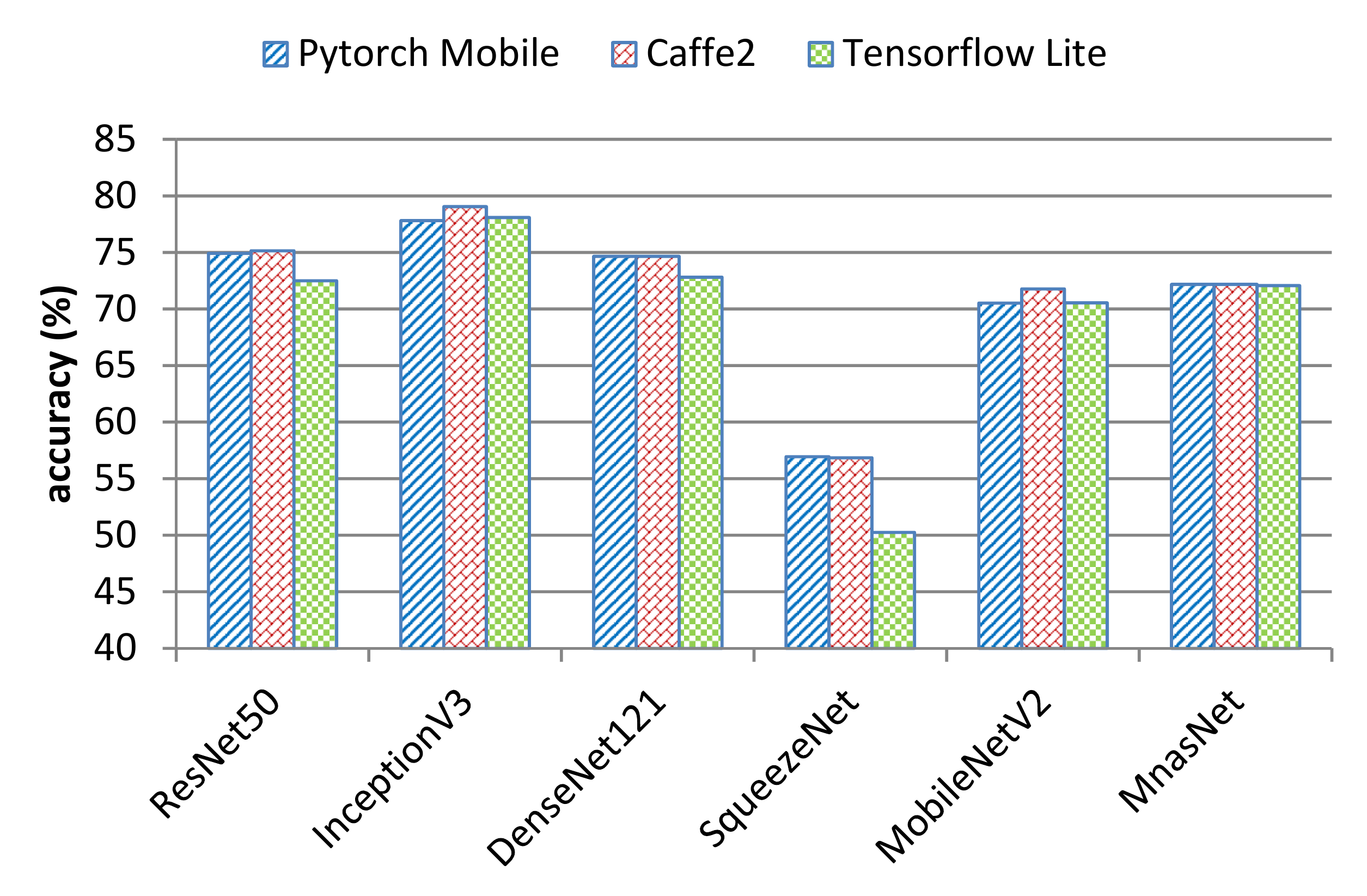}
\caption{The accuracies (\%) of different models with different frameworks.  The results are the same on the devices of Galaxy s10e, Honor v20, Vivo x27 and Vivo nex. The results on Oppo R17 are slightly different.}
\label{fig-acc}
\end{figure}

The same model with different implementation have different accuracy, as shown in Fig.~\ref{fig-acc}.
Because of the diverse sources, the detail of the training procedure of each model source may differ with each other.  For example,  they may have different strategies of data preprocessing, optimization. Also, they may have different hyper-parameters, e.g. batch size, or parameter initialization.

Moreover,  the same model with same implementation have different accuracy on different devices. From the Table \ref{table-result},  we can see that the same model with same implementation achieves the same accuracy on Galaxy s10e, Honor v20, Vivo x27 and Vivo nex. However, it has slightly different accuracy on Oppo R17.

\textbf{Remark 2: There is no one-size-fits-all solution for AI frameworks on mobile devices.}

\begin{figure}[tbh!]
\centering
\includegraphics[width=1\columnwidth]{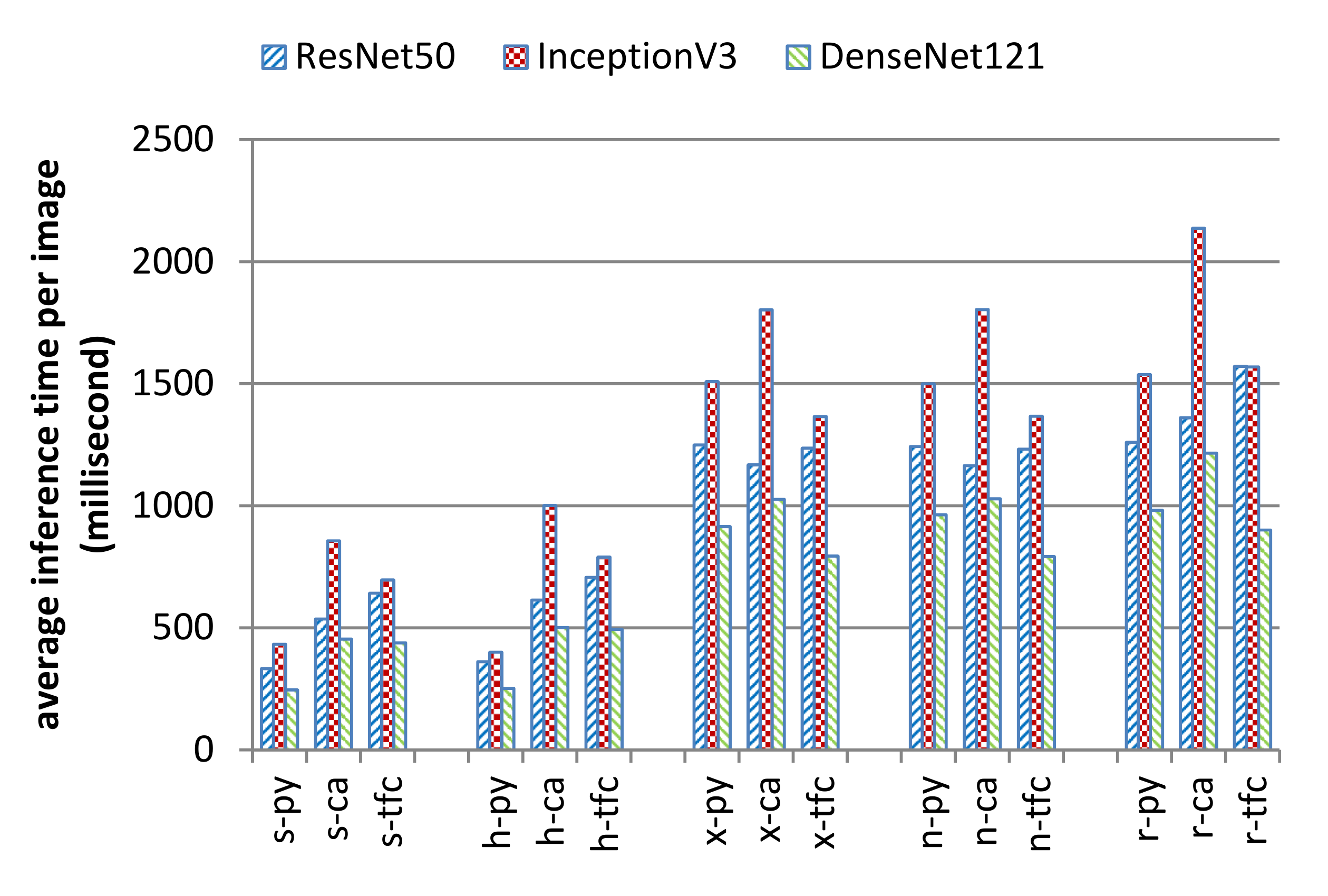}
\caption{The average inference time of the heavy-weight networks with different frameworks. (s: Galaxy s10e, h: Honor v20, x: Vivo x27, n: Vivo nex, r: Oppo R17. py: Pytorch Mobile, ca: Caffe2, tfc: Tensorflow Lite with CPU. ) }
\label{fig-heavy}
\end{figure}

\begin{figure}[tbh!]
\centering
\includegraphics[width=1\columnwidth]{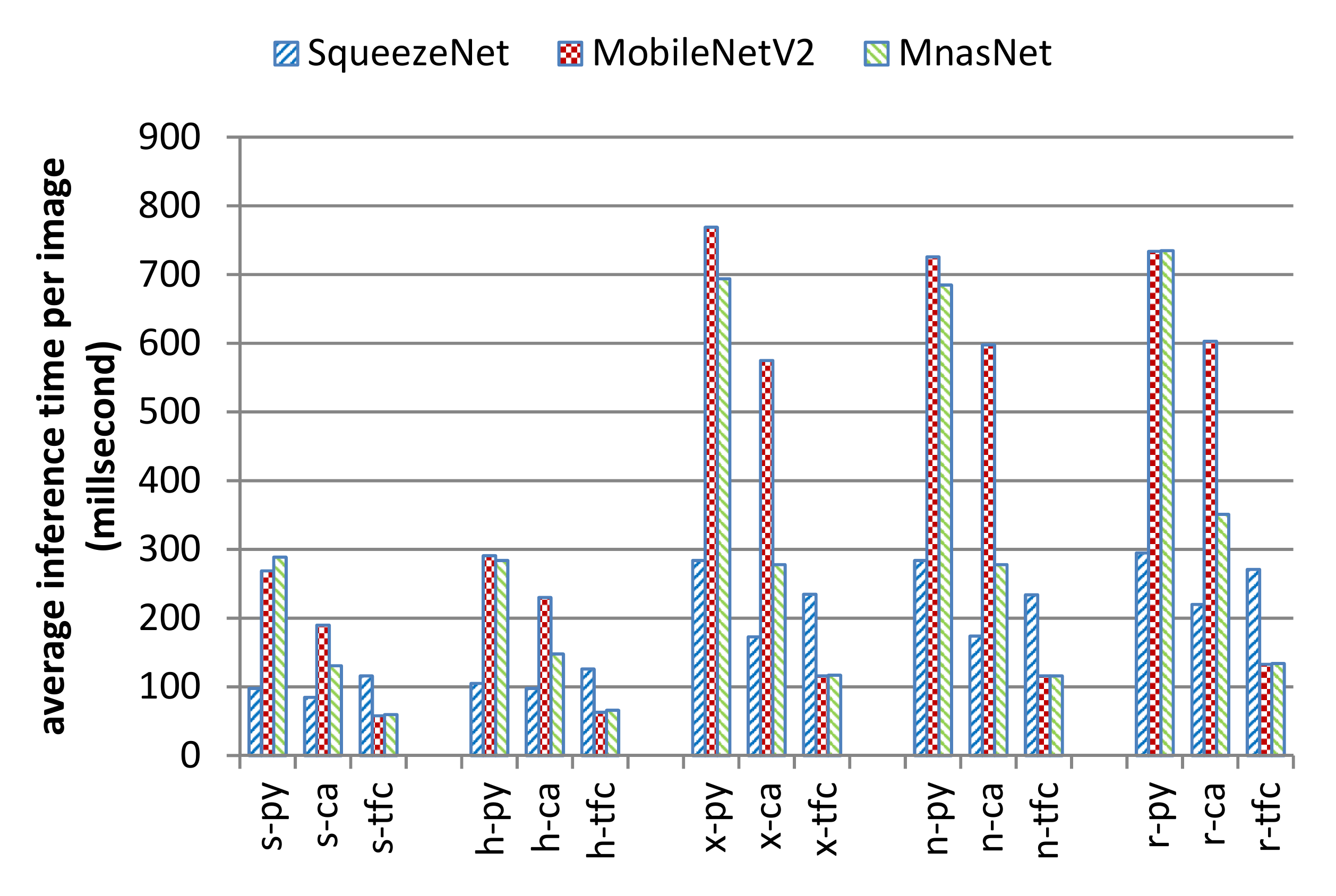}
\caption{The average inference time of the light-weight networks with different frameworks. (s: Galaxy s10e, h: Honor v20, x: Vivo x27, n: Vivo nex, r: Oppo R17. py: Pytorch Mobile, ca: Caffe2, tfc: Tensorflow Lite with CPU. )  }
\label{fig-light}
\end{figure}

For convenience, we shows the average inference time of different test on Fig.~\ref{fig-heavy} and \ref{fig-light}, and we compare the performance of the model with Pytorch Mobile, Caffe2 and Tensorflow Lite CPU on different devices. Depending on both the model and the device, different framework has different performance.

On Galaxy s10e and Honor v20, for all the three heavy networks, implementations with Pytorch Mobile perform better than Caffe2 and Tensorflow Lite. For ResNet50, Caffe2 performs better than Tensorflow Lite CPU, while for InceptionV3, Caffe2 performs worse than Tensorflow Lite CPU. 

On Vivo x27 and Vivo nex, for ResNet50, Caffe2 implementation performs best, and Tensorflow Lite CPU performs better than Pytorch Mobile.  For InceptionV3 and DenseNet121, Tensorflow Lite CPU performs best, Caffe2 performs worst. 

On Oppo R17, for ResNet50, the implementation with Pytorch Mobile perform best, and Caffe2 performs better than Tensorflow Lite CPU. For InceptionV3, Pytorch Mobile and Tensorflow Lite CPU achieve similar performances,  better than Caffe2. For DenseNet121, Tensorflow Lite CPU performs best,  and Pytorch Mobile performs better than Caffe2.

On all the measured device, for  MoibleNetV2 and MnasNet, Tensorflow Lite performs best, and Pytorch Mobile performs worst. For SqueezeNet, Caffe2 performs best on all the measured device,  while  Tensorflow Lite CPU performs worst on  Galaxy s10e and Honor v20, Pytorch Mobile performs worst on Vivo x27 and Vivo nex and Oppo R17.

\textbf{Remark 3: NNAPI does not always accelerate the inference.}

The Android Neural Networks API (NNAPI), available after Android 8.1, aims to accelerate for AI models on Android devices with supported hardware accelerators including: Graphics Processing Unit (GPU), Digital Signal Processor (DSP), Neural Processing Unit (NPU). 
It is advised that developers do not use NNAPI directly in applications, but instead use higher-level machine learning frameworks which uses NNAPI to perform hardware-accelerated inference operations on supported devices \cite{nnapi}.

\begin{figure}[tbh!]
\centering
\includegraphics[width=1\columnwidth]{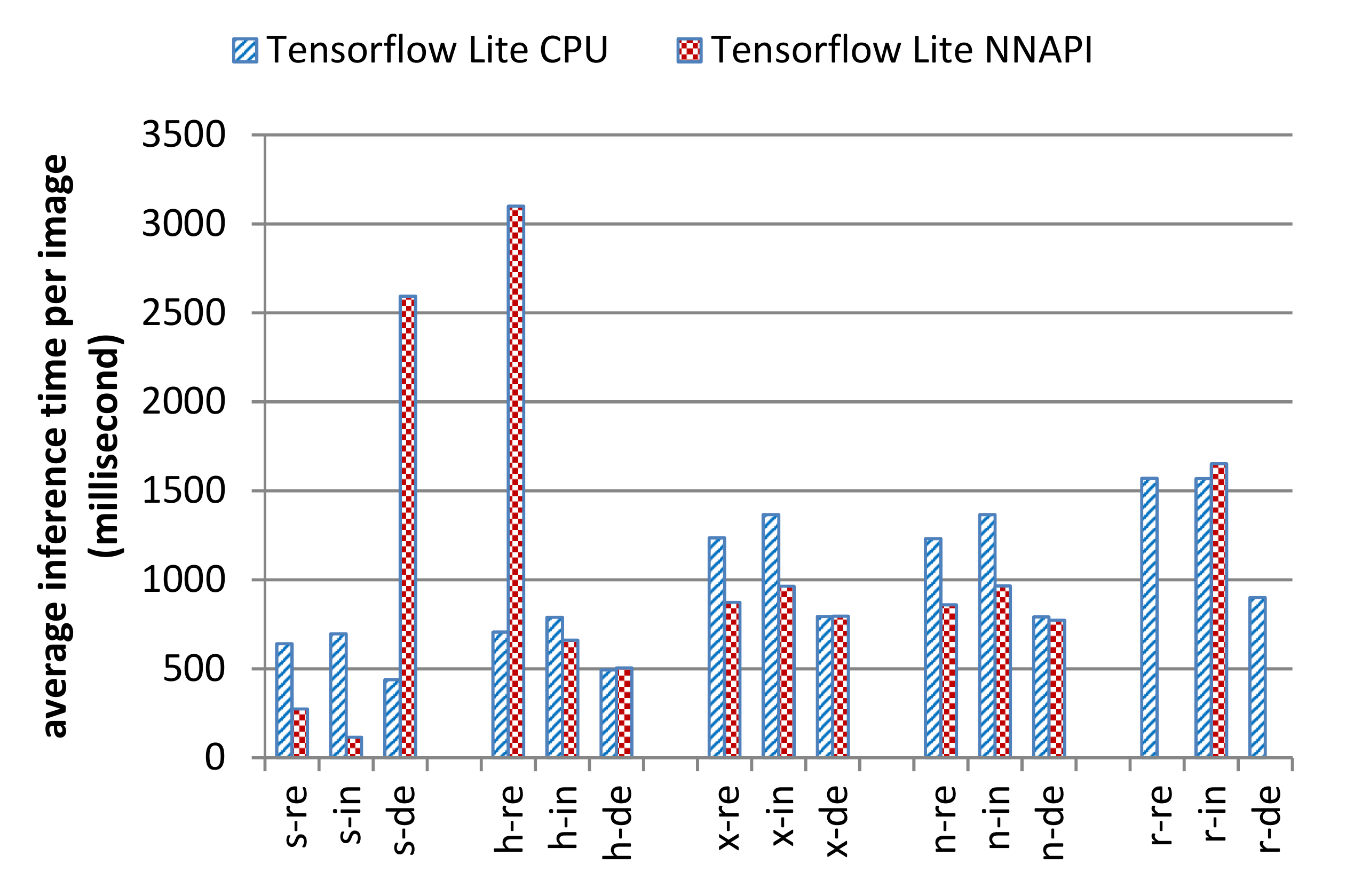}
\caption{The average inference time of the heavy-weight networks with and without NNAPI delegate. (s: Galaxy s10e, h: Honor v20, x: Vivo x27, n: Vivo nex, r: Oppo R17.  re: ResNet50, in: InceptionV3, de: DenseNet121, sq: SqueezeNet, mo: MobileNetV2, mn: MnasNet. ) }
\label{fig-heavy-tf}
\end{figure}

\begin{figure}[tbh!]
\centering
\includegraphics[width=1\columnwidth]{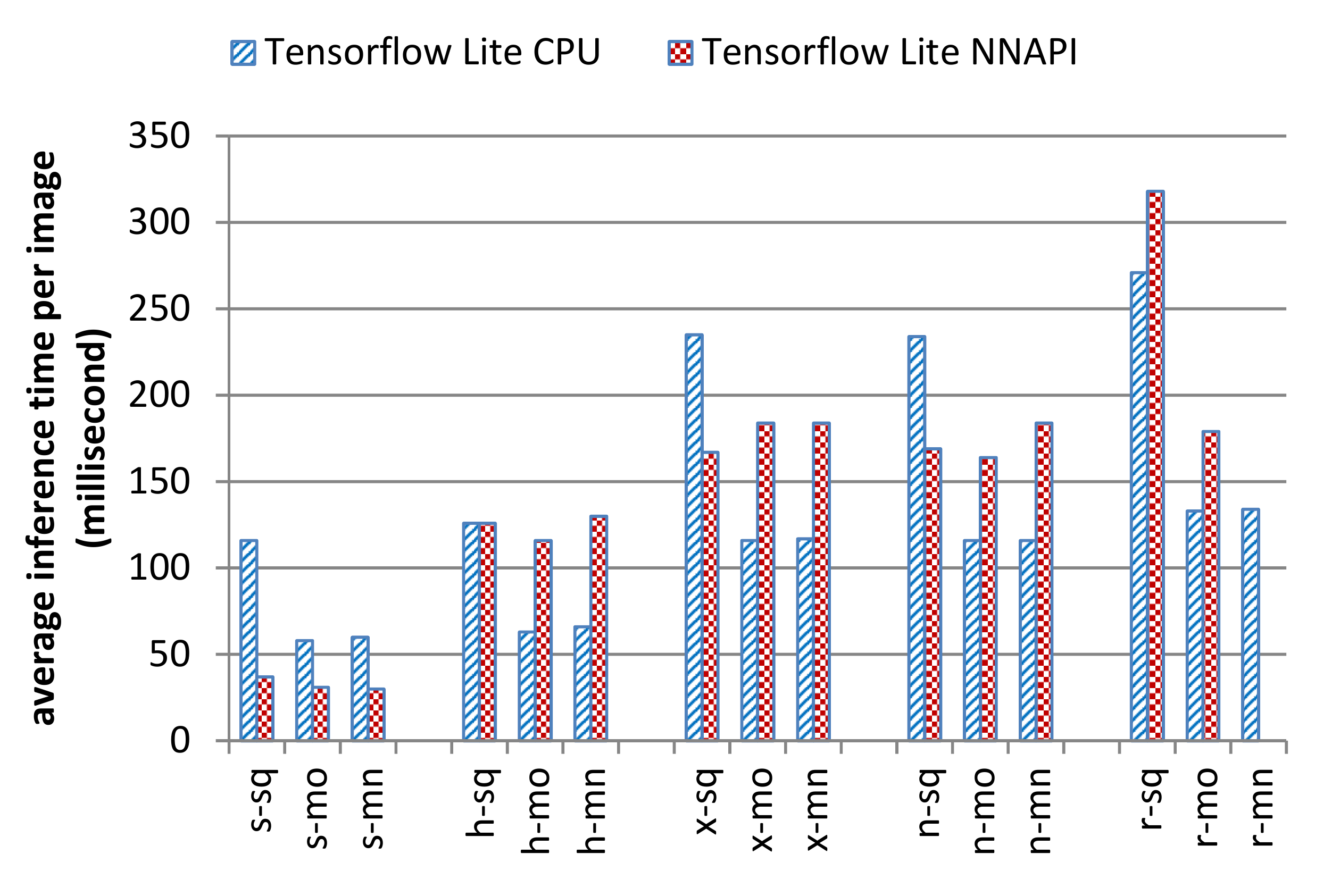}
\caption{The average inference time of the light-weight networks with and without NNAPI delegate. (s: Galaxy s10e, h: Honor v20, x: Vivo x27, n: Vivo nex, r: Oppo R17.  re: ResNet50, in: InceptionV3, de: DenseNet121, sq: SqueezeNet, mo: MobileNetV2, mn: MnasNet.)  }
\label{fig-light-tf}
\end{figure}

Currently, only Tensorflow Lite support the NNAPI delegate. 
However, NNAPI does not always accelerate the inference comparing without NNAPI delegate.  
For heavy-weight networks, as shown in Fig.~\ref{fig-heavy-tf}, NNAPI  dramatically slows down the DenseNet121 on Galaxy s10e. Also, it dramatically slows down the ResNet50 on Honor v20. On Oppo R17, with NNAPI, ResNet50 and DenseNet121 do not work at all, and InceptionV3 is slowed down. In other cases of heavy-weight networks, NNAPI do accelerate the inference. As shown in Fig.~\ref{fig-light-tf}, NNAPI  accelerates the inference of all the three light-weight networks on the Galaxy s10e,  SqueezeNet on Vivo x27 and Vivo nex. However, it  slows down the inference in all other cases, and does not work for the MnasNet on Oppo R17.

The reason is that not all the operators are supported by the NNAPI delegate. 
Tensorflow Lite first checks the model to decide whether the operators in the model are supported by the delegate. If there are operators that are not supported, the model graph is partitioned into several sub-graphs, and the unsupported sub-graphs are ran on the CPU. This brings an overhead of communication between the CPU and delegate \cite{Lee2019OnDeviceNN}.

\textbf{Remark 4: Model complexity does affect the inference time, but the implementation is also important.}

Model complexity refers to the computational complexity (FLOPs) and the number of parameters. Actually, the architecture heterogeneity of the network also could reflect the complexity, since the  heterogeneity makes the optimization and parallelism more difficult.  Theoretically, as mentioned in Section \ref{sec-model}, InceptionV3 has the most FLOPs and parameters, and it is most heterogeneous among the three heavy-weight networks, while DenseNet121 has the fewest  FLOPs and parameters. 
Among the three light-weight networks, SqueezeNet has the most FLOPs and fewest parameters, while MnasNet has the most FLOPs and parameters, and it is most heterogeneous.

There is no doubt that the light-weight networks are more efficient than heavy-weight networks. 
Among the heavy-weight networks, as shown in Fig.~\ref{fig-heavy}, with the same framework, InceptionV3 does take more inference time, and DenseNet121 takes less inference time. However, with different framework, less complexity does not always bring less inference time. For example, on Galaxy s10e and Honor v20, DenseNet121 with Caffe2 and Tensorflow Lite CPU takes more time than ResNet50 with Pytorch Mobile. Among the light-weight networks, as shown in Fig.~\ref{fig-light}, SqueezeNet takes less time than MobileNetV2 and MnasNet with Pytorch Mobile and Caffe2. However, it takes more time than MobileNetV2 and MnasNet with Tensorflow Lite CPU. For MobileNetV2 and MnasNet, MnasNet is more efficient with Pytorch Mobile and Caffe2 (except Pytorch Mobile on Galaxy s10e),  and they have close performances with Tensorflow Lite CPU.  The influence of the implementation is also reflected in Tensorflow Lite NNAPI.  For example,  with NNAPI, InceptionV3 is more efficient than ResNet50, and ResNet50 is more efficient than DenseNet121 on Galaxy s10e.

\textbf{Remark 5:  Tensorflow Lite takes the least time to load the model,  and Caffe2 takes the most time to load the model.}

\begin{figure}[tbh!]
\centering
\includegraphics[width=1\columnwidth]{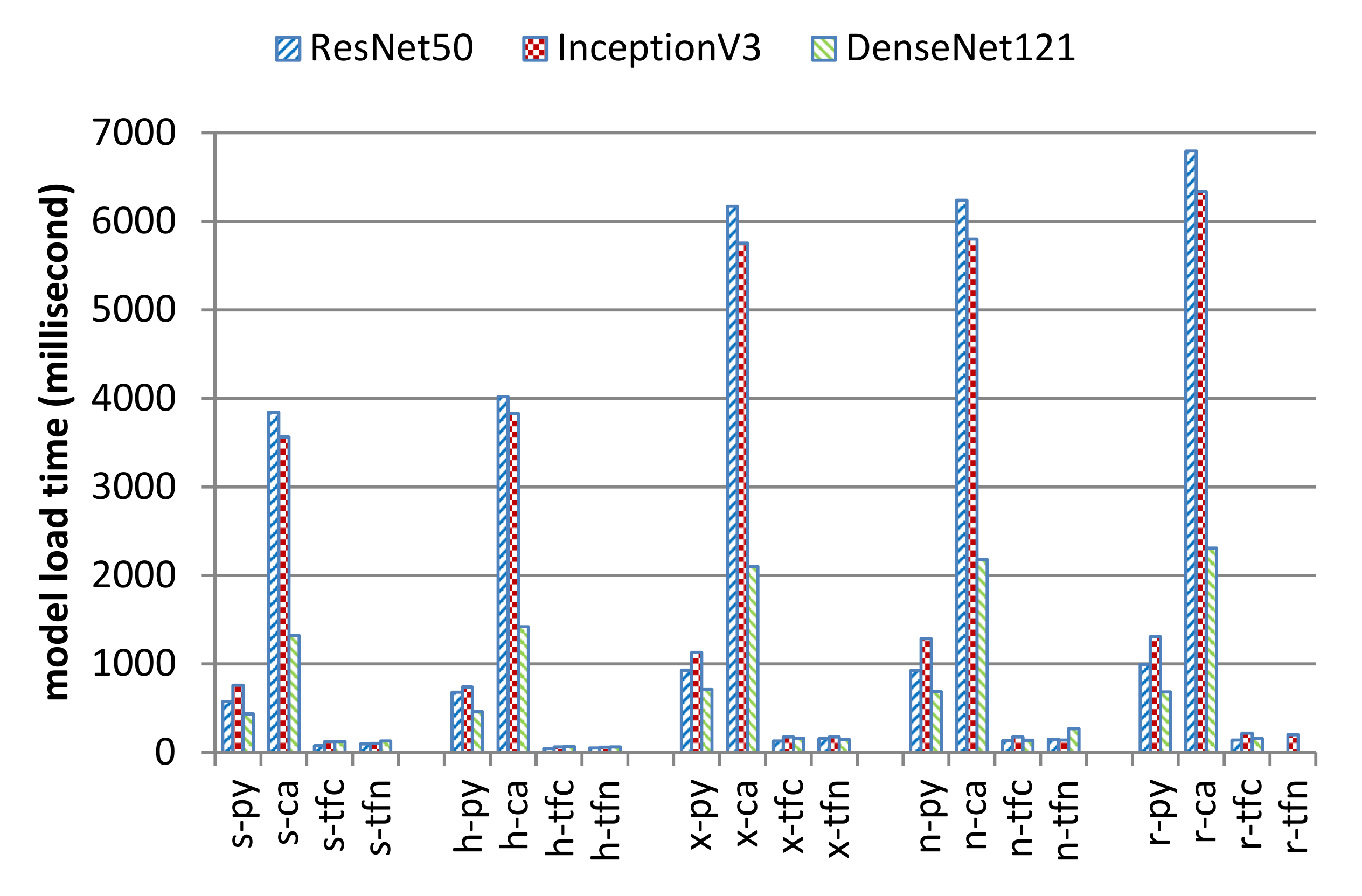}
\caption{The model load time of the heavy-weight networks with different frameworks. (s: Galaxy s10e, h: Honor v20, x: Vivo x27, n: Vivo nex, r: Oppo R17. py: Pytorch Mobile, ca: Caffe2, tfc: Tensorflow Lite with CPU, tfn: Tensorflow Lite with NNAPI delegate. ) }
\label{fig-heavy-load}
\end{figure}

\begin{figure}[tbh!]
\centering
\includegraphics[width=1\columnwidth]{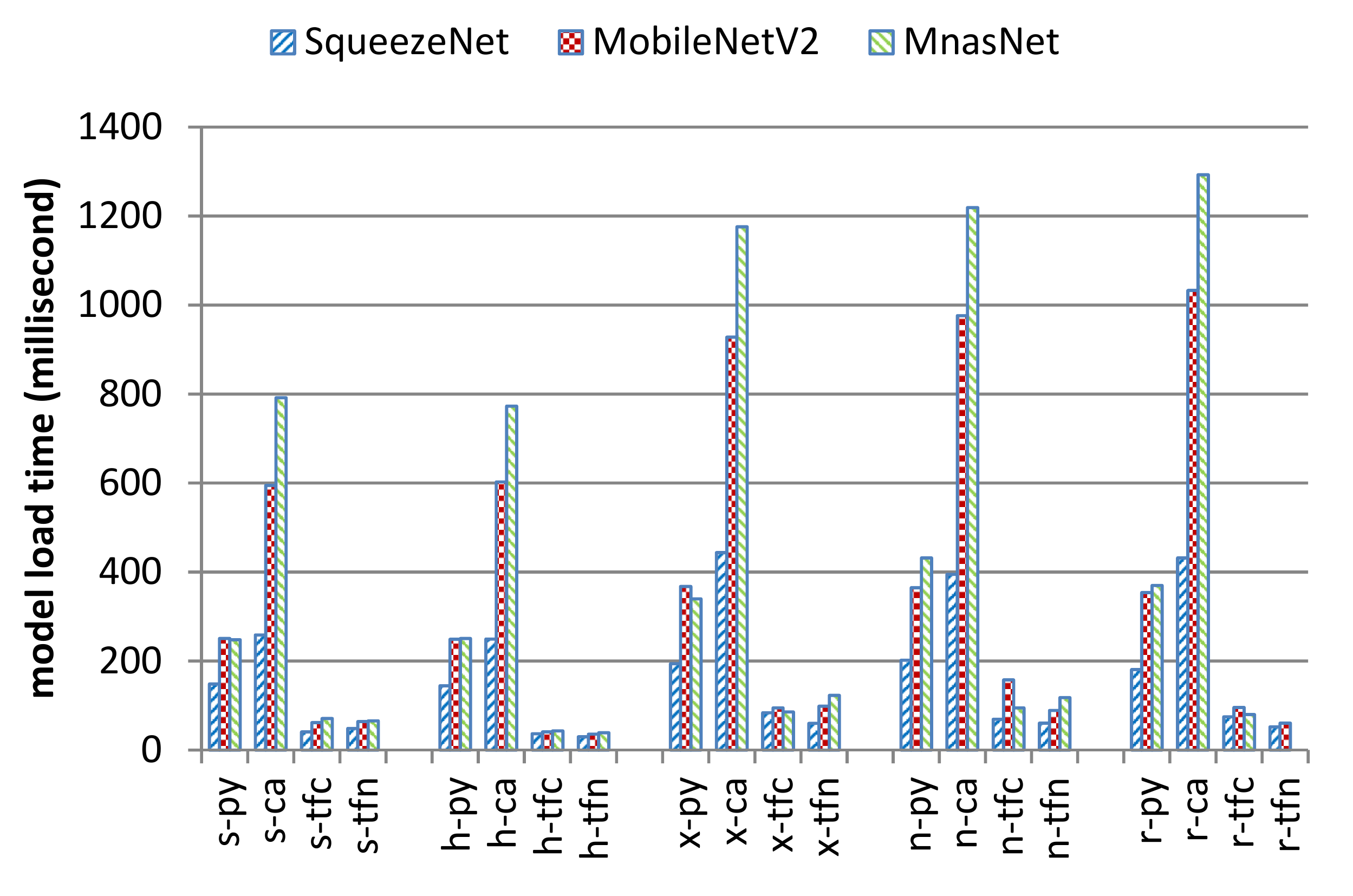}
\caption{The model load time of the light-weight networks with different frameworks. (s: Galaxy s10e, h: Honor v20, x: Vivo x27, n: Vivo nex, r: Oppo R17. py: Pytorch Mobile, ca: Caffe2, tfc: Tensorflow Lite with CPU, tfn: Tensorflow Lite with NNAPI delegate. )  }
\label{fig-light-load}
\end{figure}

Before running the inference, the model need be loaded and initialized.  Fig.~\ref{fig-heavy-load} and \ref{fig-light-load} show the load time of each model with different framework implementation. 
Tensorflow Lite use FlatBuffers to serializes the model. 
FlatBuffers, originally created at Google for game development and other performance-critical applications,  is an efficient cross platform serialization library \cite{flatbuffers}.
By using FlatBuffers, Tensorflow Lite takes the least time to load the model. 
For Caffe2, there are two model files: a init\_net.pb file stores the weights, and a predict\_net.pb file stores the operations. Additionally, Caffe2 has larger file size for the same model, and takes the most time to load the model.

\begin{table*}[tbh!]
  \caption{The accuracy (\%), the  average inference time (millisecond), and the final AI score  on different devices. (py: Pytorch Mobile, ca: Caffe2, tfc: Tensorflow Lite with CPU, tfn: Tensorflow Lite with NNAPI delegate. re: ResNet50, in: InceptionV3, de: DenseNet121, sq: SqueezeNet, mo: MobileNetV2, mn: MnasNet. Slash means the device does not support the corresponding implementation)  }
\centering
\begin{tabular}{|c|cc|cc|cc|cc|cc|}
\hline
   &  \multicolumn{2}{c|}{\textbf{Galaxy s10e}}   &  \multicolumn{2}{c|}{\textbf{Honor v20}} & \multicolumn{2}{c|}{\textbf{Vivo x27}} & \multicolumn{2}{c|}{\textbf{Vivo nex}} & \multicolumn{2}{c|}{\textbf{Oppo R17}} \\
\cline{2-11}
&accuracy&time&accuracy&time&accuracy&time&accuracy&time&accuracy&time\\
\hline
py-re   & 74.94& 333 & 74.94 &361& 74.94 & 1249&74.94 & 1243 &75.16 &1260\\
py-in   & 77.82 & 433 & 77.82 & 401& 77.82 & 1509 &77.82 &1500 &77.68&1537\\
py-de   & 74.66& 246   & 74.66& 253 & 74.66 & 915&74.66  &963&74.72&982\\
py-sq   & 56.94 & 98 & 56.94 & 105& 56.94 & 284&56.94 &284  &56.74&295\\
py-mo   & 70.54  & 269 & 70.54& 291 & 70.54& 769 &70.54 &726 &70.44&734\\
py-mn   & 72.18 & 289  & 72.18 & 284 & 72.18 & 694 &72.18  &685 &72.24&735\\

ca-re   & 75.16 & 537  & 75.16 & 614& 75.16 & 1168&75.16  &1165 &75.14&1361\\
ca-in   & 79.06 & 857 & 79.06& 1002  & 79.06 & 1803&79.06  &1804&78.9 &2137\\
ca-de   & 74.66 & 454   & 74.66 & 501 & 74.66 & 1027 &74.66 &1029 &74.72&1216\\
ca-sq   & 56.86 & 85  & 56.86 & 98 & 56.86 & 173&56.86 &174  &56.54&220\\
ca-mo   & 71.78 & 190 & 71.78& 230  & 71.78 & 575&71.78 &598 &71.64&603\\
ca-mn   & 72.18 & 131   & 72.18 & 148& 72.18 & 278&72.18  &278 &72.24&351\\

tfc-re  & 72.5 & 642   & 72.5 & 707 & 72.5 & 1236  &72.5 &1232    &72.44 &1571\\
tfc-in  & 78.1 & 697  & 78.1 & 790& 78.1  & 1366 &78.1 &1367   &78.2 &1569\\
tfc-de  & 72.82  & 439  & 72.82  & 494& 72.82 & 794  &72.82 &792 &72.78 &901\\
tfc-sq  & 50.26 & 116 & 50.26 & 126& 50.26 & 235 &50.26 &234  &50.44&271\\
tfc-mo  & 70.56& 58    & 70.56 & 63 & 70.56& 116  &70.56 &116 &70.58 &133\\
tfc-mn  & 72.08 & 60   & 72.08 & 66 & 72.08 & 117  &72.08  &116&72.08 &134\\

tfn-re  & 72.5 & 275  & 72.5 & 3100 &72.5  & 874 &72.5   &860&/ &/\\
tfn-in  & 78.1  & 116  & 78.1  & 661&78.1 & 965  &78.1  &966  &78.2&1653\\
tfn-de  & 72.82  & 2595 & 72.82& 505 &72.82 & 796 &72.82&773  &/&/\\
tfn-sq  & 50.26  & 37   & 50.26 & 126 &50.26 & 167 &50.26 &169  &50.44&318\\\
tfn-mo  & 70.56 & 31  & 70.56  & 116 &70.56 & 184  &70.56  &164  &70.58 &179\\
tfn-mn  & 72.08& 30   & 72.08& 130 &72.08 & 184 &72.08  &184 &/  &/  \\
\hline
 \textbf{AI score (VIPS)}   &  \multicolumn{2}{c|}{140.40 }   &  \multicolumn{2}{c|}{82.73} & \multicolumn{2}{c|}{44.61 } & \multicolumn{2}{c|}{45.11} & \multicolumn{2}{c|}{33.40} \\
\hline
 \textbf{AI score (VOPS)}   &  \multicolumn{2}{c|}{151.19G }   &  \multicolumn{2}{c|}{92.79G } & \multicolumn{2}{c|}{47.87G } & \multicolumn{2}{c|}{48.05G } & \multicolumn{2}{c|}{34.15G} \\
\hline
  \end{tabular}
  \label{table-result}
\end{table*}

\textbf{Remark 6: The AI scores enable a direct comparison and ranking of different devices .}

The details of the accuracy, the average inference time and the final AI score are shown in Table \ref{table-result}.  Galaxy s10e achieves the highest AI scores, and Honor v20 achieves the second highest AI scores. Galaxy s10e and Honor v20  use more advanced SoCs. Comparing to Galaxy s10e, Honor v20 does not well support for Tensorflow Lite NNAPI.  The configurations of Vivo x27 and Vivo nex are similar, and they achieve the close scores.  Oppo R17 uses lowest Soc, and achieves the lowest AI scores.

\section{Conclusions and Future Works}
In this paper, we propose a benchmark suite, AIoTBench, which focuses on the evaluation of the inference on mobile or embedded devices. The workloads in our benchmark cover more model architectures and more frameworks. Specifically, AIoTBench covers three typical heavy-weight networks: ResNet50, InceptionV3, DenseNet121, as well as three light-weight networks: SqueezeNet , MobileNetV2, MnasNet. Each model is implemented  by three frameworks: Tensorflow Lite, Caffe2, Pytorch Mobile. Our benchmark is more available and affordable for the users, since it is off the shelf and needs no re-implementation. We analyze the diversity and the representativeness of the selected models and frameworks, and describe the technical detail of the benchmarking. 
Moreover, we propose two unified metrics as the AI scores: Valid Images Per Second (VIPS) and Valid FLOPs Per Second (VOPS). They reflect the trade-off between quality and performance of the AI system.

Currently, we have compared and ranked 5 mobile devices using our benchmark. This list will be extended and updated soon after. The current version of AIoTBench focuses on the vision session. In the future, we will extend our benchmark to other sessions, e.g. language session. Moreover, we will apply our benchmark on other mobile and embeded devices in addition to smartphones. We will also consider the model quantization in our future works.

\bibliographystyle{IEEEtran}
\bibliography{references}

\end{document}